\documentclass[lettersize,journal]{IEEEtran}
\usepackage{amsmath,amsfonts}
\usepackage{algorithmic}
\usepackage{algorithm}
\usepackage{array}
\usepackage[caption=false,font=normalsize,labelfont=sf,textfont=sf]{subfig}
\usepackage{makecell}
\usepackage{textcomp}
\usepackage{stfloats}
\usepackage{url}
\usepackage{verbatim}
\usepackage{graphicx}
\usepackage{cite}
\usepackage{times}
\usepackage{epsfig}
\usepackage{graphicx}
\usepackage{amsmath}
\usepackage{amssymb}
\usepackage{algorithm,algorithmic}
\usepackage{multirow}
\usepackage{subfloat}
\usepackage{newfloat}
\usepackage{listings}
\usepackage{color}
\usepackage{lineno}
\usepackage{bibentry}
\usepackage{bbding}

\begin{document}

\title{UCDFormer: Unsupervised Change Detection Using a Transformer-driven Image Translation}

\author{Qingsong Xu,  Yilei Shi,~\IEEEmembership{Member,~IEEE}, Jianhua Guo, Chaojun Ouyang, Xiao Xiang Zhu,~\IEEEmembership{Fellow,~IEEE}
\thanks{The work is jointly supported by the German Research Foundation (DFG GZ: ZH 498/18-1; Project number: 519016653), by the German Federal Ministry of Education and Research (BMBF) in the framework of the international future AI lab ``AI4EO -- Artificial Intelligence for Earth Observation: Reasoning, Uncertainties, Ethics and Beyond" (grant number: 01DD20001) and by German Federal Ministry for Economic Affairs and Climate Action in the framework of the ``national center of excellence ML4Earth" (grant number: 50EE2201C) and by TUM in the framework of TUM Innovation Network ``EarthCare: Twin Earth Methodologies for Biodiversity, Natural Hazards, and Urbanisation". (Corresponding author: Xiao Xiang Zhu)}
\thanks{Q. Xu, J. Guo and X. Zhu are with the Chair of Data Science in Earth Observation (SiPEO), Technical University of Munich (TUM), 80333 Munich, Germany. (e-mails: qingsong.xu@tum.de; jianhua.guo@tum.de; xiaoxiang.zhu@tum.de).
}
\thanks{Y. Shi is with the Chair of Remote Sensing Technology, Technical University of Munich (TUM), 80333 Munich, Germany. (e-mail: yilei.shi@tum.de).}
\thanks{C. Ouyang is with the Key Laboratory of Mountain Hazards and Surface Process, Institute of Mountain Hazards and Environment, Chinese Academy of Sciences, Chengdu 610041, China (e-mail: cjouyang@imde.ac.cn).}}

\markboth{IEEE TRANSACTIONS ON GEOSCIENCE AND REMOTE SENSING}%
{Shell \MakeLowercase{\textit{et al.}}: A Sample Article Using IEEEtran.cls for IEEE Journals}


\maketitle

\begin{abstract}
\textcolor{blue}{This work has been accepted by IEEE TGRS for
publication.} Change detection (CD) by comparing two bi-temporal images is a crucial task in remote sensing. With the advantages of requiring no cumbersome labeled change information, unsupervised CD has attracted extensive attention in the community. However,  existing unsupervised CD approaches rarely consider the seasonal  and style differences incurred by the illumination and atmospheric conditions in multi-temporal images. To this end, we propose a change detection with domain shift setting for remote sensing images. Furthermore, we present a novel  unsupervised CD method using a light-weight transformer, called UCDFormer. 
 Specifically,  a transformer-driven image translation composed of a light-weight transformer and a domain-specific affinity weight is first proposed to mitigate domain shift between two images with real-time efficiency. 
 After image translation, we can generate the difference map between the translated before-event image and the original after-event image. 
 Then, a novel reliable pixel extraction module is proposed to select significantly changed/unchanged pixel positions by fusing the pseudo change maps of fuzzy c-means clustering and adaptive threshold. Finally, a binary change map is obtained based on these selected pixel pairs and a binary classifier.
 Experimental results on different unsupervised CD tasks with seasonal  and style changes demonstrate the effectiveness of the proposed UCDFormer. For example, compared with several other related methods, UCDFormer improves performance on the Kappa coefficient by more than 12\%. In addition, UCDFormer achieves excellent performance for   earthquake-induced landslide detection when considering large-scale applications.
 The code is available at \textcolor{red}{\url{https://github.com/zhu-xlab/UCDFormer}}.
 
\end{abstract}

\begin{IEEEkeywords}
Change detection, Domain shift, Unsupervised Learning, Transformer, UCDFormer.
\end{IEEEkeywords}

\section{Introduction}
\IEEEPARstart{C}{hange} detection (CD) from remote sensing images is a technique to understand the changed information of the same objects observed at different times. Real-time and accurate access to surface changes is of great importance for Earth observation applications, such as environmental evolution~\cite{guo2023nationwide}, urbanization~\cite{wang2022graph, zhan2020unsupervised}, land-use monitoring~\cite{zhu2014continuous, zhu2017}, resource management~\cite{kennedy2009remote}, and disaster assessment~\cite{zheng2021building, xu2022mffenet}.
Many CD methods have been proposed in recent years~\cite{shafique2022deep,wen2021change, 9069898, liu2021abnet}. According to deep learning-based change extraction techniques, current CD methods can be divided into two categories: supervised methods and unsupervised methods. Supervised methods train a CD model by large amounts of labeled remote sensing data~\cite{toker2022dynamicearthnet,mou2018learning,wang2018getnet,wang2021novel,luo2023multiscale}. Recently, supervised CD models based on vision transformer~\cite{chen2021remote,dosovitskiy2020image,wang2022hybrid, zhang2023asymmetric} have achieved outstanding performance. Specifically, a deep multi-scale Siamese architecture, called
MSPSNet~\cite{guo2021deep} has been proposed to extract multi-scale features by using a parallel convolutional structure and self-attention. A Swin transformer-based network, called SwinSUNet~\cite{zhang2022swinsunet} has been proposed for supervised CD.  The SwinSUNet contains three modules: encoder, a fusion module, and decoder. Multi-scale features are obtained by employing a hierarchical Swin transformer~\cite{liu2021swin}. In addition,  a hybrid architecture, TransUNetCD~\cite{li2022transunetcd}, is presented to obtain effective  multi-scale features by combining the merits of transformers and UNet. A transformer-based scene change detection architecture (TransCD)~\cite{wang2021transcd} is proposed to establish global semantic relations and model long-range context by incorporating a siamese vision transformer.
The main challenge of these  supervised CD methods is the design of an effective module to obtain global-local context features within the spatial and temporal
scope  in identifying the change of interest in remote sensing images. However, these transformer-based CD methods rely on manually labeling training samples, which 
prevents the context modeling advantages of transformer from being exploited in unsupervised CD. 

In contrast, unsupervised methods can extract changes automatically, and thus are more popular in practical applications~\cite{noh2022unsupervised,luppino2021deep,saha2019unsupervised,9627707}. These methods are generally composed of three parts.

1$)$ Building effective representation of  multi-temporal images at the pixel-level~\cite{liu2021unsupervised,luppino2021deep,kalinicheva2020unsupervised}, feature-level~\cite{saha2019unsupervised,wu2021unsupervised,de2019unsupervised}, or object-level~\cite{zheng2021unsupervised,zhang2021style}. Importantly, homogeneous data captured by the same types of sensors and heterogeneous images captured by different types of sensors use different methods to establish an effective representation of  multi-temporal images. Specifically, spatial-temporal features can be obtained directly by deep convolutional neural networks or graph neural networks for  homogeneous data, such as multi-scale feature convolution unit~\cite{chen2019change}, deep change vector (DCV)~\cite{saha2019unsupervised}, deep Siamese KPCA convolutional mapping network~\cite{wu2021unsupervised}, temporal-spatial-structural graph~\cite{wu2023unsupervised}, and fully convolutional change detection framework with generative adversarial network~\cite{wu2023fully}. In addition, for heterogeneous images, the direct extraction of representation between heterogeneous images is difficult because different sensors may capture different statistical properties~\cite{liu2021unsupervised}. Thus, an image-to-image translation network is designed to translate one image into the pixel space of another one and thereby transfer the issue as CD of the homogeneous images based on generative adversarial networks (GANs), such as the cycle-consistent adversarial networks (CycleGANs) \cite{liu2021unsupervised, luppino2021deep}, and conditional generative adversarial network (cGAN)~\cite{niu2018conditional}. 

2$)$ Calculating change differences by comparing the extracted representation of multitemporal images. Computationally efficient Euclidean distance is usually used to obtain change differences. 

3$)$ Generating the final change map by setting appropriate thresholds or clustering methods. Examples include Otsu’s global thresholding or local adaptive thresholding in DCVA~\cite{saha2019unsupervised}, k-means clustering in~\cite{celik2009unsupervised}, and the fuzzy c-means (FCM) clustering algorithm in DSMS-FCN~\cite{chen2019change}. 
\begin{figure*}[t]
	\centering
	{\includegraphics[width = .9\textwidth]{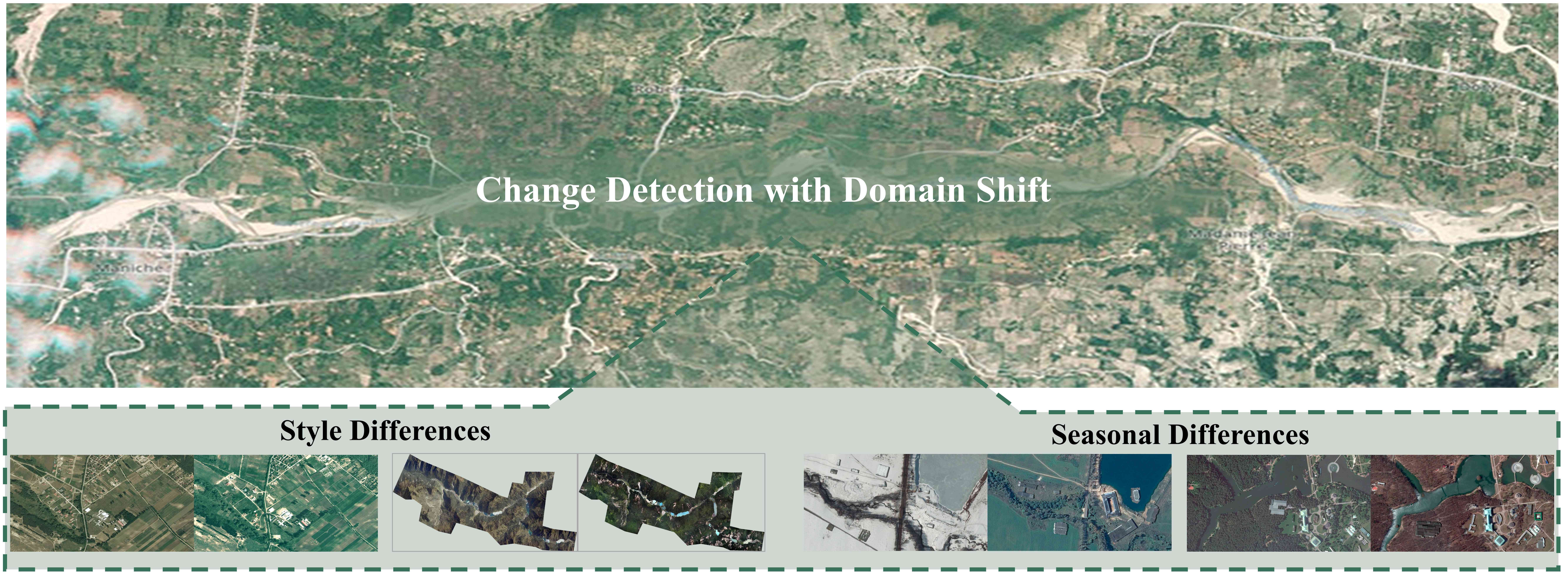}}
	\caption{\small{Change detection with domain shift, which is used to reflect style differences and seasonal differences between multi-temporal images.}}
	\label{fig:0}
\end{figure*}

The main difficulties affecting unsupervised CD in remote sensing images are  differences in light conditions, atmospheric conditions, and seasonality due to different acquisition dates. 
However,  these unsupervised CD methods have the following bottlenecks when addressing this problem.

\begin{list}{\labelitemi}{\leftmargin=10pt \topsep=0pt \parsep=0pt}
	\item  Existing unsupervised CD approaches rarely consider the seasonal differences and variations incurred by the illumination and atmospheric conditions in multi-temporal images.
	\item GAN-based image translation methods~\cite{zhu2017unpaired} for heterogeneous images  will change the content of objects and backgrounds during the translation process. In addition, they can only get a definite output according to the input. Once the input and output are changed, the GAN-based model needs to be retrained. Arbitrary attribute transfer cannot be realized by GAN-based image translation methods. Thus,  it is necessary to develop an image translation method for homogeneous images.
	\item  Most unsupervised CD methods for calculating change differences only use low-level features, failing to mine the deep information of multi-temporal images. Furthermore, they are based on direct comparison (such as generating the final change map by setting thresholds or clustering methods). There may be some noise and accumulation of errors in the final result.
\end{list}

To deal with these challenges, a scenario of \textbf{change detection with domain shift} has been proposed.  The concept of domain shift~\cite{ben2010theory} comes from domain adaptation~\cite{9262039, xu2023universal}, which is used to reflect the distribution differences between two domains. As shown in Fig.~\ref{fig:0}, the setting of change detection with domain shift is used to detect changes, considering style differences and seasonal differences between multi-temporal images.

Bearing these concerns in mind, in this paper, we try to provide an unsupervised CD using a light-weight transformer called UCDFormer, by defining a three-step approach: 1$)$ a transformer-driven image translation module is proposed to map data across two domains with real-time efficiency, which first leverages the contextual modeling benefits of transformer in the unsupervised CD of remote sensing images; 2$)$ a reliable pixel extraction module is proposed to extract significantly changed/unchanged pixel positions from the difference map, which is computed by fusing multi-scale feature maps; and 3$)$ a binary change map is obtained based on the reliable changed/unchanged pixel positions and the random forest (RF) classifier~\cite{pal2005random}. More specifically,  a light-weight transformer is proposed in the transformer-driven image translation module to reduce  the computational complexity of  the self-attention layer in regular transformers~\cite{vaswani2017attention, dosovitskiy2020image,chen2021remote} by using spatial downsampling and channel group operations, inspired by StyleFormer~\cite{wu2021styleformer} and GG-Transformer~\cite{yu2021glance}. Importantly, a comparison of domain-specific affinity weight is proposed to compute a priori change indicator in an unsupervised manner. The affinity weight is intended to  reduce the translation strength of changed pixels and to increase the translation of unchanged pixels, by using a weighted translation loss.  Compared with some classical GAN-based unsupervised image translation approaches (such as CycleGANs~\cite{liu2021unsupervised, luppino2021deep}, cGAN~\cite{niu2018conditional}, SCAN~\cite{li2018unsupervised}, and DTCDN~\cite{li2021deep}), our proposed transformer-driven image translation module can produce visually plausible stylization results for arbitrary attribute transfer with real-time efficiency, and the content coherence with the input content images.
In addition, the proposed  affinity weight is an efficient approach that provides more useful information than these GAN-based image translation methods. Thanks to the proposed transformer-driven image translation module, the domain shift due to style differences and  seasonal differences for homogeneous data can be significantly reduced in image space. Furthermore, to overcome the third bottleneck of existing unsupervised CD methods, a novel reliable pixel extraction module is proposed that fuses the results of FCM clustering and adaptive threshold.  The FCM clustering and adaptive threshold methods are complementary, as the former can avoid noise in the final change map and  the interference caused by the image translation, and the latter can avoid error accumulation in the final change map. Finally, based on the reliable changed and unchanged pixel positions, the corresponding pixels are extracted from bi-temporary images and are then used to learn a binary classifier. The other pixel pairs will be classified by this classifier to obtain the final CD results.

In a nutshell, our contributions are listed as follows.
\begin{list}{\labelitemi}{\leftmargin=10pt \topsep=0pt \parsep=0pt}
	\item We introduce a change detection with domain shift setting for remote sensing images. In order to reduce the domain shift in image space, a novel unsupervised CD method using a light-weight transformer, called UCDFormer, is proposed.
	\item A transformer-driven image translation module in UCDFormer is proposed to map data across two domains with real-time efficiency, by using efficient self-attention.
	\item An affinity weight is proposed to compute a measure of similarity between bi-temporal images with domain shift.
	\item An advanced reliable pixel extraction approach is presented to produce significant changed and unchanged pixel pairs for a binary classifier.
\end{list}

The rest of this paper is organized as follows.  Section~\ref{me} elaborates the proposed UCDFormer. Section~\ref{ex} provides experimental settings, experimental results, and discussion. Finally, Section~\ref{co} draws the conclusion of our work in this paper.


\section{Methodology \label{me}}

\subsection{Transformer-driven Image Translation}
\begin{figure*}[t]
	\centering
	{\includegraphics[width = .95\textwidth]{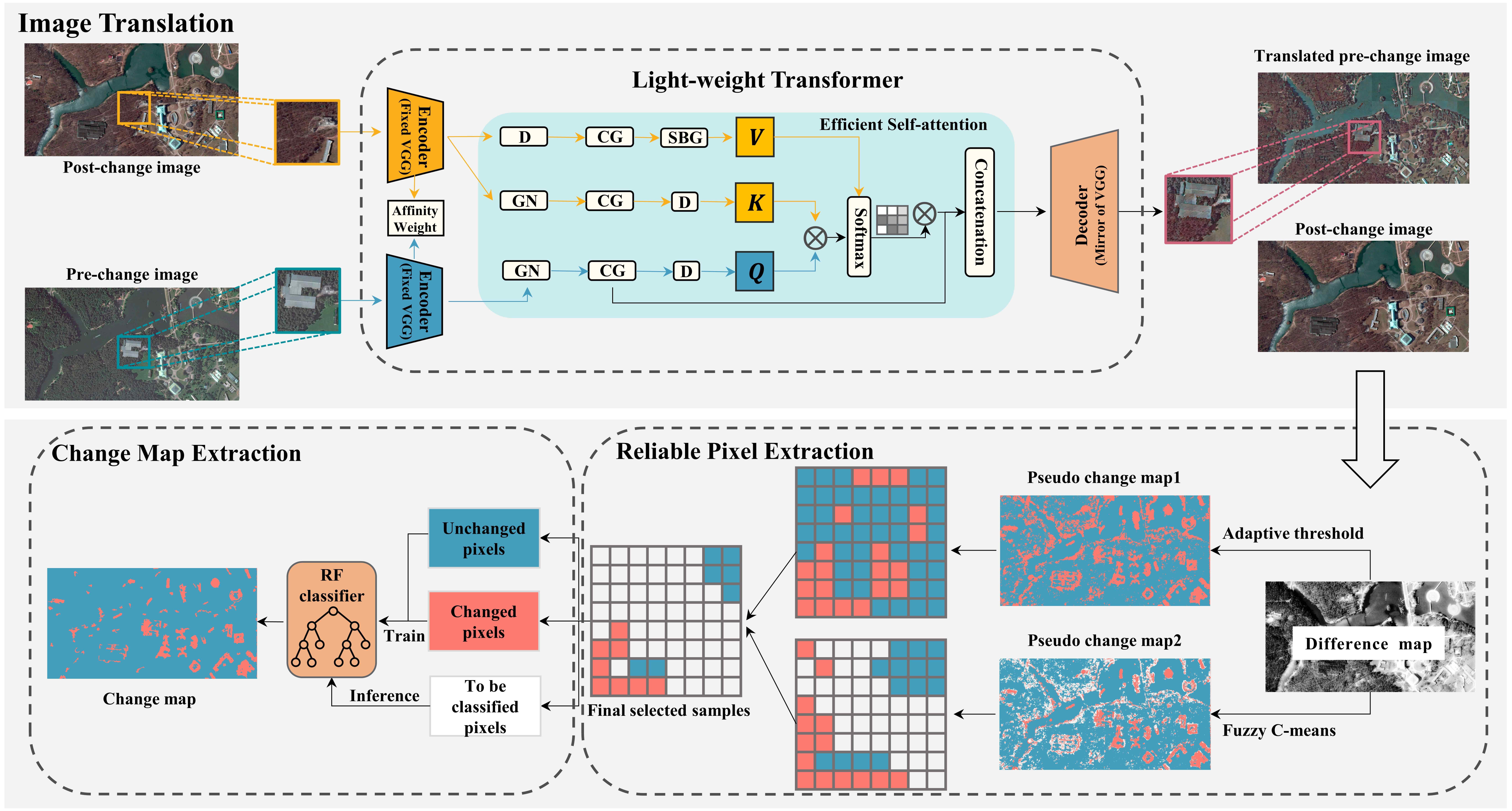}}
	\caption{\small{Overview of UCDFormer.  The architecture of the model is divided into three parts, 	i.e., image translation, reliable pixel extraction, and change map extraction.  D, GN, CG, SBG represent downsampling, group normalization, channel group, and style bank generation, respectively.}}
	\label{fig:1}
\end{figure*}

In order to address the task of CD from multi-temporal imagery with domain shift (such as seasonal differences and style differences), a transformer-driven image translation module is proposed to map data between two images with real-time efficiency.
Unlike GAN-based image-to-image translation~\cite{liu2021unsupervised, zhang2021style,luppino2021deep}, transformer-driven image translation adopts the expressive multi-head attention strategy from the well-known transformer architecture~\cite{vaswani2017attention} to globally model a new image, whose image content is coherent with the pre-change image, and whose style is optionally the same as the post-change image. As Fig.~\ref{fig:1} illustrates, this entails training a transformer-driven function $F(I_1, I_2): I_1, I_2 \rightarrow \hat{I}$ to translate bi-temporal remote sensing images ($I_1, I_2$)  with domain shift to distinguishable homogeneous remote sensing images ($\hat{I}, I_2$) for change map extraction, where 
 $I_1$ and $I_2$ are the bi-temporal remote sensing images at two different time instants, $T_1$ and $T_2$. 
 
Specifically, a fixed-weight encoder $E$ (a 16-layer VGG network pretrained on the
ImageNet~\cite{simonyan2014very}) is first used to obtain the representative feature maps $Z_{pre} \in \mathbb{R}^{H \times W \times C}$ and $Z_{pos} \in \mathbb{R}^{H \times W \times C}$ from $I_1$ and $I_2$, respectively, where 
 $H$ and $W$
 are the height and the width of the pre-change and post-change feature maps, respectively, and $C$ is the number of channels. The $Z_{pre}$  is then transferred to the translated feature $\hat{Z}$ according to $Z_{pos}$, based on a proposed light-weight transformer. Finally, the translated feature $\hat{Z}$ is fed into a learnable decoder $D_{\boldsymbol{\theta}}$. 
 
\subsubsection{A Light-weight Transformer}
The main computation bottleneck of the regular transformer~\cite{vaswani2017attention, dosovitskiy2020image} is the self-attention layer. In the original multi-head self-attention process, each of the heads $Q$, $K$, $V$ have the same dimensions $N \times C$, where $N = H \times W$ is the length of the sequence, and the self-attention is estimated as
\begin{equation}
\operatorname{Attention}(Q, K, V)=\operatorname{Softmax}\left(\frac{Q K^{\top}}{\sqrt{d_{h e a d}}}\right) V.
\label{eq1}
\end{equation}

The computational complexity of this process is $O\left(N^2\right)$, which is prohibitive for large-scale remote sensing images. Inspired by StyleFormer~\cite{wu2021styleformer} and GG-Transformer~\cite{yu2021glance}, an efficient self-attention is proposed by using spatial downsampling ($D$) and channel group (CG) operations.

\textit{a) Efficient self-attention:} The channels of feature maps $Z_{pre} \in \mathbb{R}^{H \times W \times C}$ and $Z_{pos} \in \mathbb{R}^{H \times W \times C}$ are first normalized group-wise by  Group Normalization (GN). Then, 
 the group-wise normalized channels are divided into $G$ groups, which can improve computational efficiency. Furthermore, the spatial size $H \times W$ is downsampled by a factor of $m$ by two convolutional layers with stride=2. Finally, the $K$, $Q$ are obtained by
 \begin{equation}
 \begin{aligned}
Q &= D(\operatorname{CG}(\operatorname{GN}(Z_{pre}))),\\
Q^{(g)} & \in \mathbb{R}^{(H / m) \times (W / m) \times (C / G) \times 1}, g \in\{1, \ldots, G\}, \\
K &= D(\operatorname{CG}(\operatorname{GN}(Z_{pos}))),\\
K^{(g)} & \in \mathbb{R}^{(H / m) \times (W / m) \times (C / G) \times 1}, g \in\{1, \ldots, G\}.
 \end{aligned}
 \end{equation}
 The style value $V$ contains ($(H / m) \times (W / m)$) values by downsampling of two-layer convolutions  with stride=2. In addition, the channel of each value is divided into $G$ groups to ensure efficiency. Importantly, to produce the  diverse style or distribution patterns,
 an  affine transformation in each group is utilized by using $J$ variants, which are generated by using the style bank generation (SBG) module. The concept of SBG is from StyleFormer~\cite{wu2021styleformer}. The fundamental principle of SBG is to generate affine transformation matrices by using channel expansion. The SBG module is re-implemented in our efficient self-attention by using  two convolution layers with a kernel of 3 and a stride of 1. Importantly, the output channel of the second convolution layer is $G \times J \times (C / G) \times(C / G+1)$. Thus, each value $V_i, i \in\left\{1, \ldots, (H / m) \times (W / m)\right\}$ has $G \times J$ affine transformation matrices with the size of $(C / G) \times(C / G+1)$. This formula is expressed as follows.
\begin{equation}
 \begin{aligned}
 V &= \operatorname{SBG}(\operatorname{CG}(D(Z_{pos}))),\\
 V^{(i)} &  \in \mathbb{R}^{ G \times J \times(C / G) \times(C / G+1)}, i \in\left\{1, \ldots, (H / m) \times (W / m)\right\}.
 \end{aligned}
 \end{equation}
 
\textit{b) Concatenation:}
After the efficient self-attention, the group-wise affine coefficients $\{\hat{A}^{g}\}_{g=1}^G, \hat{A}^{g} \in \mathbb{R}^{(H / m) \times (W / m)  \times J \times (C / G) \times (C / G+1)}$ is obtained.   Then two corresponding group-wise affine transformation matrices, ${W}^{(g)} \in \mathbb{R}^{(H / m) \times (W / m) \times (C / G) \times (C / G)}$ and ${b}^{(g)} \in \mathbb{R}^{(H / m) \times (W / m) \times (C / G) \times 1}$, are sampled from $J$ affine transformation matrices in $\{\hat{A}^{g}\}$ by using the slicing operation~\cite{wu2021styleformer}. The selected affine coefficients are then applied to transform the group-wise normalized $Q^{(g)}$ into the group-wise feature representation $\hat{Z}^{g}$, which is represented as
 
 \begin{equation}
 \hat{Z}^{g}= {W}^{(g)} \otimes \hat{A}^{g}+{b}^{(g)},
 \end{equation}
 where $\otimes$ indicates the matrix multiplication operation at every spatial position. The final translated feature $\hat{Z} \in \mathbb{R}^{H \times W  \times C}$ can be obtained by concatenating $\{\hat{Z}^{g}\}_{g=1}^G$ along the channel dimension and upsampling from spatial size $(H / m) \times (W / m)$ to spatial size $H \times W$.
 
\subsubsection{Affinity Weight}
A main challenge in the image translation module is to increase the intra-class compactness and inter-class distinguishability of changed and non-changed classes, respectively. To manage this challenge effectively, an affinity weight is proposed to compute a measure of similarity between bi-temporal images with domain shift. The goal of the affinity weight is to reduce the translation strength of changed pixels  and to increase the translation of unchanged pixels. The  affinity weight $\alpha$ computed from pre-change image $I_1$ and post-change image $I_2$ is depicted in Algorithm~\ref{alg:1}. Note that the procedure is totally unsupervised and does not require any ancillary information or knowledge about the data.

A $k \times k$ non-overlapping sliding window covers an area $p$ and an area $q$ of $I_1$ and $I_2$, respectively, from which a pair of corresponding patches $p_{\ell}$ and $q_{\ell}$ ($\ell \in\{1, \ldots,|\mathcal{P}|\}$, where $P$ is the number of patches)  are firstly extracted. Then, multi-scale feature maps $E(p)$ and $E(q)$ from each layer are obtained by inputting $p_{\ell}$ and $q_{\ell}$ into the fixed-weight encoder. Multi-scale feature maps $E(p)$ and $E(q)$ are then concatenated to obtain deep feature vectors $V_{p} \in \mathbb{R}^{k \times k \times C_{V}}$ and $V_{q} \in \mathbb{R}^{k \times k \times C_{V}}$  (where $C_{V}$ is the channel of deep feature vectors), respectively. Assuming that unchanged pixels yield similarly deep hypervectors~\cite{saha2019unsupervised}, the computationally efficient Euclidean distance can be used to describe the difference between changed and unchanged areas. Thus, the distances between all pixel pairs from $V_{p}$ and $V_{q}$ are computed by
\begin{equation}
d_{\ell}(V_{p}, V_{q})=\frac{1}{C_{V}}\sum_{i=1}^{C_{V}} \left\|V_{p}-V_{q}\right\|_2^2, d_{\ell} \in \mathbb{R}^{k \times k}. 
\end{equation}
Once computed, the patch-wise affinity weight $\alpha_{\ell} \in \mathbb{R}^{k \times k}$ is computed by $1 - \operatorname{normalized}(d_{\ell})$. Finally, the pixel-wise affinity weight $\alpha \in \mathbb{R}^{H \times W}$ is obtained by concatenating patch-wise affinity weights.

\begin{center}
	\begin{algorithm}[htbp!]
		\caption{Calculation of Affinity Weight $\alpha$:}
		\begin{algorithmic}[1]
			\FOR{all patches $p_{\ell} \in \mathbb{R}^{k \times k}$ from $I_1$, $q_{\ell} \in \mathbb{R}^{k \times k}$ from $I_2$, $\ell \in\{1, \ldots,|\mathcal{P}|\}$}
			\STATE Obtain multi-scale feature maps $E(p)$ and $E(q)$ by using the fixed-weight encoder
			\STATE Obtain deep feature vectors $V_{p} \in \mathbb{R}^{k \times k \times C_{V}}$ and $V_{q} \in \mathbb{R}^{k \times k \times C_{V}}$ by concatenating multi-scale feature maps
			\STATE Compute $d_{\ell}(V_{p}, V_{q})=\frac{1}{C_{V}}\sum_{i=1}^{C_{V}} \left\|V_{p}-V_{q}\right\|_2^2$ and normalize $d_{\ell}$
			\STATE Compute $\alpha_{\ell} = 1 - d_{\ell}$, $\alpha_{\ell} \in \mathbb{R}^{k \times k}$
			\STATE Add $\alpha_{\ell}$ to the affinity set $S$
			\ENDFOR
			\FOR{affinity set $S$}
			\STATE Obtain $\alpha \in \mathbb{R}^{H \times W}$ by concatenating $S$
			\ENDFOR
		\end{algorithmic}
		\label{alg:1}
	\end{algorithm}
	\vspace{-5mm}
\end{center}
\subsubsection{Optimization}
The proposed transformer-driven image translation network is trained in a supervised fashion similar to the style transfer methods~\cite{johnson2016perceptual, wu2021styleformer}. The training losses consist of a weighted translation loss and a content loss. 

\textit{a) Weighted translation loss:} 
The weighted translation loss $\mathcal{L}_{wt}$ matches the mean and standard deviations of the VGG-16 features between the translated image and the weighted style image. This is
\begin{align}
\mathcal{L}_{wt}=&\sum_{l}\left\|\mu_{l}\left(\hat{I}\right)-\mu_{l}({I_w})\right\|_{2}^{2}  \\
&+\sum_{l}\left\|\sigma_{l}^{2}\left(\hat{I}\right)-\sigma_{l}^{2}({I_w})\right\|_{2}^{2}, 
\end{align}
where $\mu_{l}\left(\hat{I}\right)$ and $\sigma_{l}^{2}(\hat{I})$ denote the batch-wise mean and variance estimates of the feature maps obtained by inputting translated image $\hat{I}$ to the fixed-weight encoder at the $l$-th layer. The parameters  $\mu_{l}(I_w)$ and $\sigma_{l}^{2}(I_w)$ are the corresponding mean and variance, derived by inputting weighted style image $I_w$ to the encoder. So that the proposed translation network's ability to discriminate changes  is not hindered. Two criteria for the  weighted style image should be satisfied: the style of unchanged areas should come from the post-change image $I_2$ and the style of changed areas should come from the pre-change image $I_1$. Thus, the weighted style image $I_w$ is obtained by affinity weight.
\begin{equation} 
I_w=\alpha \otimes I_2 + (1 - \alpha) \otimes I_1.
\end{equation}
Specifically,  when the pixel is likely changed, $\alpha$ is supposed to be close to zero, and $I_w$ comes from  $I_1$. Conversely, when the pixel is likely unchanged,  $\alpha$ is close to one, and $I_w$ comes from  $I_2$.

\textit{b) Content loss:}  
The content loss $\mathcal{L}_{c}$ matches the self-similarity features extracted from the fixed-weight encoder between the translated image and the pre-change image. It is represented as
\begin{equation}
\textstyle \mathcal{L}_{c}=\sum_{l}\left\|E_{l}\left(\hat{I}\right)-E_{l}(I_1)\right\|_{2}^{2}, 
\end{equation}
where $E_{l}\left(\hat{I}\right)$ and $E_{l}(I_1)$ denote the feature maps of $\hat{I}$ and $I_1$ at the $l$-th layer, respectively. The semantic information and spatial layout can be efficiently preserved by $\mathcal{L}_{c}$.

Thus, we train a transformer-driven image translation network by minimizing
\begin{equation}
\mathcal{L}= \mathcal{L}_{wt}+\gamma \mathcal{L}_c,
\label{eq.10}
\end{equation}
where $\gamma$ is a hyperparameter for balancing the loss, and $\gamma$ is set to 1000 by default in our experiments.

\subsection{Change Extraction}
The multi-scale feature maps of the translated image $\hat{I}$ and the post-change image $I_2$ can be obtained by inputting $\hat{I}$ and $I_2$ into the fixed-weight encoder, respectively. Then, the deep feature vectors  of $\hat{I}$ and $I_2$ are computed by fusing multi-scale feature maps. 
The pixel-wise difference map $D$ is obtained by computing the Euclidean distance between the deep feature vectors. 
Numerous experiments have demonstrated that difference maps obtained through feature space are more representative and distinguishable than those obtained in image space. The difference map $D$ can help to detect the change/nonchange region directly by using clustering or threshold methods. However, it cannot be used as the final result, because the translated image $\hat{I}$ depends on the reliability of transformer-driven image translation. If the binary change detection result is obtained directly through thresholding or clustering methods, there may be some noise and accumulation of errors in the final result. Therefore, a reliable pixel extraction module is proposed to extract significantly changed/unchanged pixel positions from the difference map $D$. 

\subsubsection{Reliable Pixel Extraction}
Reliable pixel extraction is achieved by fusing the results of   FCM clustering and adaptive threshold.

\textit{a) FCM clustering:} The FCM clustering algorithm~\cite{chen2019change,gao2016automatic,fang2022novel} is based on the memberships of pixels in the difference map. Specifically, FCM clustering is performed to cluster $D$ into three categories: $\omega_c$, $\omega_u$, and $\omega_n$. Two of these,  $\omega_c$ and $\omega_n$, are reliable pixels that have high change and unchange probabilities, respectively. The third, $\omega_u$ represents unreliable pixels and needs to be classified. 
The  intensity centers of the three clusters are represented as $c_1$, $c_2$, and $c_3$ ($c_1<c_2<c_3$), and their corresponding cluster categories are $k_1$, $k_2$, and $k_3$. A pseudo-change map can be obtained from these values  using the following rule,
\begin{equation}
d(i, j) \in\left\{\begin{array}{cc}
\omega_c, & \text { if } FCM(d(i, j))=k_3 \\
\omega_u, & \text { if } FCM(d(i, j))=k_2 \\
\omega_n, & \text { if } FCM(d(i, j))=k_1,
\end{array}\right.
\label{eq.11}
\end{equation}
where $d(i, j)$ denotes the pixel value of the $i$-th row and $j$-th column in $D$. The predicted class by the FMC clustering algorithm is denoted by $FCM(d(i, j))$.

\textit{b) Adaptive threshold:} Any suitable thresholding method can be employed to determine a binary change map. In this work, the adaptive threshold method based on  Gaussian filtering is used to decide the change/unchange region. The adaptive threshold is a popular threshold method for binary change detection~\cite{saha2019unsupervised}. Based on this method, the decision boundary value $\mathcal{T}$ is applied to the difference map $D$. Thus, another pseudo-change map is generated based on the following rule,
\begin{equation}
d(i, j) \in \begin{cases}\omega_c, & \text { if } d(i, j) \geq \mathcal{T}\\ 
\omega_n, & \text { if } d(i, j) < \mathcal{T}. \end{cases}
\label{eq.12}
\end{equation}

\textit{c) Extraction of reliable changed and unchanged pixels:} 
The FCM clustering and adaptive threshold methods are complementary, as the former can avoid noise in the final change map and  the interference caused by the image translation, and the latter can avoid error accumulation in the final change map. In order to integrate their advantages to generate more reliable samples, the fusion strategy shown in Fig.~\ref{fig:1} is utilized. Specifically, if a pixel is classified as the changed class in both pseudo-change map $P_1$ generated by FCM clustering and pseudo-change map $P_2$ generated by the adaptive threshold, it is considered to be a changed sample. Conversely, a pixel is considered an unchanged sample if it is regarded as the unchanged class in both pseudo-change maps. Furthermore, to-be-classified samples are equal to the total pixels of $D$ minus the determined changed samples and the determined unchanged samples. This is  expressed as
\begin{equation}
d(i, j) \in \begin{cases}
\omega_c, &  \text { if } P_1(d(i, j))=P_2(d(i, j))=\omega_c \\
\omega_n, &  \text { if } P_1(d(i, j))=P_2(d(i, j))=\omega_n \\
\omega_u, &  \text { Otherwise. } \end{cases}
\end{equation}
\subsubsection{Change Map Extraction}
Based on the reliable changed and unchanged pixel positions obtained from reliable pixel extraction, we can quickly extract the corresponding pixels from the pre-change image $I_1$ and the post-change image $I_2$. In this way, the CD with domain shift is transformed into a classical binary classification problem. In such a case, the core contains three elements: the training set $R_r$, the test set $R_e$, and the classifier $\Gamma(\cdot)$. The training set $R_r$ contains two types of information: changed or unchanged pixel pairs by stacking  $I_1$ and $I_2$, and the reliable pseudo-label $\omega_c$ and $\omega_n$. The test set $R_e$ is composed of to-be-classified pixel pairs from  $I_1$ and $I_2$, based on the positions of $\omega_u$. The RF classifier~\cite{pal2005random} is used as $\Gamma(\cdot)$ because it usually produces accuracy and robust classification performance with superior stability. Of course, other basic classifiers (such as a light-weight CNN~\cite{fang2022novel} or a Siamese CNN~\cite{chen2019change}) can also be used for change map extraction. Thus, the RF classifier is trained using the training set $R_r$. The trained classifier is then utilized to classify the pixel pairs of $R_e$ into changed or unchanged classes. The process is denoted as
\begin{equation}
P_{i}=\Gamma\left(i \in R_e \mid R_r\right),
\end{equation}
where $P_{i}$ is the classification result of to-be-classified pixels. 
By improving the result of to-be-classified pixels, we can reduce the noisy influence on the final change map. Finally, the proposed change map extraction method is robust for the difference map.

\section{Experiments and Results  \label{ex}}

\subsection{Datasets}
We evaluate the proposed network on three datasets: data with seasonal differences, data with style differences, and earthquake-induced landslide detection with style differences. Each is described below.

\subsubsection{Data with Seasonal Differences}
The dataset~\cite{zhao2020using} has three bands  with a spatial resolution of 0.3 m  that describe two different seasonal variations; the size of the images is $2700 \times 4275$ pixels. A large number of change targets have been ``increased'' and ``decreased'' within the image pairs in different seasons, and are used to test  the proposed model for its ability to detect seasonal differences. The first seasonal difference (Fig.~\ref{fig:21}) and the second seasonal difference (Fig.~\ref{fig:22}) are from summer to autumn and from spring to winter, respectively.

\subsubsection{Data with Style Differences} A image pair with a size of $ 952 \times 640$ is from  SZTAKI Air Change Benchmark~\cite{benedek2009change}. The resolution of the image pair is 1.5 m/pixel. As shown in Fig.~\ref{fig:3}, the image pair contains a pair of preliminary registered
input images and a mask of the relevant changes, which is used to test the proposed model for its ability to detect style differences.

\subsubsection{Earthquake-induced Landslide Detection with Style Differences} A magnitude $M_S$ 7.0 earthquake occurred in Jiuzhaigou County, Sichuan Province, China on August 8, 2017.  Numerous earthquake-induced landslides 
caused at least 29 road obstructions and damage in the scenic
area. As shown in Fig.~\ref{fig:4}, the research region chosen in
the Jiuzhaigou Valley covers an area of 53.6 $km^2$~\cite{xu2022mffenet}. The pre- and post-earthquake remote sensing images with a size of $4658 \times 10282$ are selected from 1.5-m resolution satellite images (Google Earth, December 2016) and 1.5-m resolution UAV images (December 2017), respectively. This earthquake-induced landslide detection can be effectively used to verify the proposed model with style differences.
\subsection{Implementation}
The proposed UCDFormer consists of two modules, transformer-driven image translation and change extraction. 
For the transformer-driven image translation, the number of channel group $G$ and the factor $m$ of downsampling are set as 16 and 4 in the efficient self-attention, respectively. The encoder $E$ is up to ReLU3$\_$1 layer in the fixed-weight VGG16~\cite{simonyan2014very}. The decoder $D_{\boldsymbol{\theta}}$(·) mirrors the encoder, as suggested by~\cite{wu2021styleformer}. There is no normalization layer in our models.
The decoder is randomly initialized at the training stage. Adam~\cite{kingma2014adam} with a learning rate of
0.0001 is used for the  transformer-driven image translation. A StepLR policy with gamma = 0.5 is used for the Adam optimizer. Furthermore, for pre-change and post-change remote sensing images, overlapping crops (overlap rate = 0.29)  are utilized to create training and test sets with patch size = $256 \times 256 \times 3$ for the transformer-driven image translation. In addition, epoch = 5000, and batch size = 1 are set for the training on different datasets.  For the change extraction, the same overlapping crop  strategy is utilized. In addition,  for the training of the RF classifier in the change extraction module,  if the difference in sample size between changed pixels and unchanged pixels is too large, the number of classes with more samples will be cropped to achieve sample balance. We use Pytorch for implementation on a high-performance computing cluster, with a Tesla P100 GPU. 
Furthermore, when processing two images with a patch size of $256 \times 256 \times 3$, the two stages of UCDFormer exhibit high floating point operations (approximately 130.8 G FLOPs) due to extensive matrix and tensor operations. However, UCDFormer exemplifies a remarkable light-weight design, characterized by a total parameter count of 19.9 M. Among these, 17.0 M parameters are trainable, such as the light-weight transformer module in UCDFormer.

\subsection{Evaluation Metrics and Competitors}
In order to accurately evaluate the effectiveness of UCDFormer, some evaluation criteria are adopted~\cite{gong2013fuzzy, mou2022detecting}. Specifically, the Kappa coefficient (Kappa), F1 score (F1), and overall accuracy (OA) are used to  evaluate overall prediction accuracy. Importantly, Kappa can better reflect the results of CD. The higher the Kappa value, the better the classification achieved. In addition, precision (the ability of a model to not label a true negative observation as positive), recall (the model's capability to find positive observations), false alarm rate (FAR, the probability of wrongly detected negatives), and missed detection rate (MDR, the probability that a model fails to infer true positive observations) are utilized to quantitatively evaluate the performance with respect to  changed class and unchanged class. Precision and recall are computed for both the changed and unchanged classes. 

We compare our network with some state-of-the-art CD algorithms, specifically, PCA~\cite{deng2008pca, celik2009unsupervised}, MAD~\cite{nielsen1998multivariate}, IR-MAD~\cite{nielsen2007regularized}, cGAN~\cite{niu2018conditional},  DCVA~\cite{saha2019unsupervised}, DSFA~\cite{du2019unsupervised},  KPCAMNet~\cite{wu2021unsupervised}, GMCD~\cite{tang2021unsupervised}, X-Net~\cite{luppino2021deep},  and Code-Aligned Autoencoder~\cite{luppino2022code}. In addition, a series of sensitivity experiments are designed to verify the effectiveness of the proposed UCDFormer. A brief introduction of these comparison methods follows.
\begin{list}{\labelitemi}{\leftmargin=10pt \topsep=0pt \parsep=0pt}
	\item  PCA~\cite{deng2008pca, celik2009unsupervised}  is developed by conducting k-means clustering on feature vectors which are extracted using principal component analysis (PCA). PCA is a transformation-based unsupervised approach with a fast computing speed.
	\item  MAD~\cite{nielsen1998multivariate} is a mainstream image transformation-based unsupervised change detection method based on the established canonical correlations analysis.
	\item  IR-MAD~\cite{nielsen2007regularized} is an extension of MAD by introducing an iteratively reweighted scheme. 
	\item cGAN~\cite{niu2018conditional}  is a conditional generative adversarial network (cGAN)-based translation network  for unsupervised CD.
	\item DCVA~\cite{saha2019unsupervised} uses a deep neural network to learn the change vectors between bitemporal images. The learned change vectors are further analyzed to detect the changed pixels by global thresholding or local adaptive thresholding.
	\item DSFA~\cite{du2019unsupervised} is a network based on SFA~\cite{wu2013slow} for unsupervised CD. Specifically, two deep networks are used to project the input data of bi-temporal images. Then, the SFA is deployed to suppress the unchanged components and
	highlight the changed components of the features.
	\item KPCAMNet~\cite{wu2021unsupervised} is a Siamese deep network built on weight-shared kernel PCA convolutions. The input of each branch is an image patch, and outputs are clustered into change and no change in an unsupervised manner.
	\item GMCD~\cite{tang2021unsupervised} is an unsupervised remote sensing CD method  based on graph convolutional network (GCN) and metric learning.  GMCD consists of a pre-trained Siamese FCN encoder and a pyramid-shaped decoder. 
	\item X-Net~\cite{luppino2021deep} is a  deep image translation network with an affinity-based change prior for unsupervised CD. The image translation network is composed of two fully convolutional networks, each dedicated to mapping the data from one domain to the other. Based on the transformed images, the difference map is obtained based on pixel-wise distance.
	\item Code-Aligned Autoencoder~\cite{luppino2022code} is an unsupervised methodology to align the code spaces of two autoencoders based on affinity information extracted from the input data.
	\item UCDFormer w/o Image Translation is the variant  that disregards  the transformer-driven image translation stage in UCDFormer.
	\item UCDFormer w/ adaptive threshold only is the variant that only uses the adaptive threshold (Eq.~\ref{eq.12}) to extract the changed area. 
	\item UCDFormer w/ FCM only is the variant that only uses the FCM clustering (Eq.~\ref{eq.11}) to obtain the final change map.	
\end{list}

\subsection{Experiments on Seasonal Changes}
\begin{table*}[htp!]
	\caption{Accuracy assessment on the dataset with seasonal change from summer to autumn. $\uparrow$ represents that a higher numerical value corresponds to a better result, while $\downarrow$ represents that a lower numerical value leads to a better result.}
	\centering
	\resizebox{1.0\textwidth}{!}
	{\begin{tabular}{cccccccccc}
			\hline
			& \multicolumn{4}{c}{Changed}                                      & \multicolumn{2}{c}{Unchanged}   &                      &                      &                         \\ \cline{2-7}
			\multirow{-2}{*}{Methods}              & Recall ($\uparrow$)        & Precision ($\uparrow$)     & MDR ($\downarrow$)           & FAR  ($\downarrow$)         & Recall  ($\uparrow$)       & Precision  ($\uparrow$)    & \multirow{-2}{*}{OA ($\uparrow$)} & \multirow{-2}{*}{F1 ($\uparrow$)} & \multirow{-2}{*}{Kappa ($\uparrow$)}  \\ \hline
			 PCA~\cite{celik2009unsupervised}                                    & 40.92          & 9.03           & 59.08          & 33.55         & 66.45          & 93.26          & 64.53                & 14.79                & 2.81                    \\
                                               MAD~\cite{nielsen1998multivariate}                                    & 27.19          & 24.39          & 72.81          & 6.86          & 93.14          & 94.02          & 88.18                & 25.72                & 19.32                   \\
                                               IR-MAD~\cite{nielsen2007regularized}                                 & 57.59          & 25.71          & 42.41          & 13.53         & 86.47          & 96.16          & 84.30                & 35.55                & 28.07                   \\
                                               cGAN~\cite{niu2018conditional}                                   & 20.56          & 19.96          & 79.44          & 6.70          & 93.30          & 93.52          & 87.82                & 20.26                & 13.67                   \\
                                               DCVA~\cite{saha2019unsupervised}                                    & 44.97          & 40.60          & 55.03          & 5.35          & 94.65          & 95.48          & 90.91                & 42.67                & 37.75                   \\
                                               DSFA~\cite{du2019unsupervised}                                    & 53.25          & 26.34          & 46.75          & 12.11         & 87.89          & 95.85          & 85.28                & 35.24                & 27.99                   \\
                                               KPCAMNet~\cite{wu2021unsupervised}                                & 54.10          & 36.15          & 45.90          & 7.77          & 92.23          & 96.11          & 89.36                & 43.34                & 37.73                   \\
                                              GMCD~\cite{tang2021unsupervised}           & 42.67          & \textbf{52.43}          & 57.33          & \textbf{3.15}          & \textbf{96.85}          & 95.41          & \textbf{92.78}               & 47.05                & 43.21                   \\
                                              X-Net~\cite{luppino2021deep}           & \textbf{59.28}          & 15.55          & \textbf{40.72}          & 26.18         & 73.82          & 95.71          & 72.73                & 24.64                & 14.45                   \\
  Code-Aligned Autoencoder~\cite{luppino2022code}              & 55.35          & 29.27          & 53.75          & 9.09          & 90.91          & 95.41          & 87.55                & 35.85                & 29.34                   \\ \hline
            UCDFormer                              & 59.10          & 50.21          & 40.90          & 4.77          & 95.23          & \textbf{96.62}          & 92.52                & \textbf{54.30}       & \textbf{50.25}          \\ \hline
\end{tabular}}
\label{Tab:Table1}
\end{table*}

\begin{table*}[htp!]
	\caption{Accuracy assessment on the dataset with seasonal change from spring to winter. $\uparrow$ represents that a higher numerical value corresponds to a better result, while $\downarrow$ represents that a lower numerical value leads to a better result.}
	\centering
	\resizebox{1.0\textwidth}{!}
	{\begin{tabular}{cccccccccc}
			\hline
			& \multicolumn{4}{c}{Changed}                                      & \multicolumn{2}{c}{Unchanged}   &                      &                      &                         \\ \cline{2-7}
			\multirow{-2}{*}{Methods}              & Recall ($\uparrow$)        & Precision ($\uparrow$)     & MDR ($\downarrow$)           & FAR  ($\downarrow$)         & Recall  ($\uparrow$)       & Precision  ($\uparrow$)    & \multirow{-2}{*}{OA ($\uparrow$)} & \multirow{-2}{*}{F1 ($\uparrow$)} & \multirow{-2}{*}{Kappa ($\uparrow$)}  \\ \hline
			PCA~\cite{celik2009unsupervised}                                   & 82.49          & 10.64          & 17.51          & 84.49         & 15.51          & 87.90          & 22.80                & 18.86                & 0.50                    \\
			MAD~\cite{nielsen1998multivariate}                                     & 18.11          & 9.79           & 81.89          & 20.37         & 79.63          & 88.85          & 72.94                & 12.71                & 1.64                    \\
			IR-MAD~\cite{nielsen2007regularized}                                 & 68.02          & 14.22          & 68.02          & 23.54         & 76.46          & 90.21          & 71.63                & 19.68                & 5.45                    \\
			cGAN~\cite{niu2018conditional}                                    & 45.61          & 14.26          & 54.39          & 33.46         & 66.54          & 90.93          & 64.26                & 21.72                & 6.18                    \\
			DCVA~\cite{saha2019unsupervised}                                   & 3.98           & \textbf{28.42}          & 96.02          & \textbf{1.22}          & \textbf{98.78}          & 89.40          & \textbf{88.47}                & 6.98                 & 4.43                    \\
			DSFA~\cite{du2019unsupervised}                                   & 22.24          & 16.65          & 77.76          & 13.59         & 86.41          & 90.11          & 79.44                & 19.04                & 7.54                    \\
			KPCAMNet~\cite{wu2021unsupervised}                                & 25.42          & 16.03          & 74.58          & 16.24         & 83.76          & 90.20          & 77.41                & 19.66                & 7.30                    \\
		 GMCD~\cite{tang2021unsupervised}            & 24.65          & 9.31           & 75.35          & 29.29         & 70.71          & 88.49          & 65.7                 & 13.52                & 2.69                    \\
			X-Net~\cite{luppino2021deep}             & \textbf{86.69} & 14.84          & \textbf{13.31} & 60.67         & 39.33          & \textbf{96.04} & 44.48                & 25.35                & 8.33                    \\
			Code-Aligned Autoencoder~\cite{luppino2022code}               & 29.25          & 9.84           & 70.75          & 32.70          & 67.30           & 88.63          & 63.16                & 14.72                & 1.85                    \\ \hline
			UCDFormer                              & 47.45          & 25.07          & 52.55          & 17.30         & 82.70          & 92.80          & 78.86                & \textbf{32.80}                & \textbf{21.65}          \\ \hline
	\end{tabular}}
\label{Tab:Table2}
\end{table*}
\begin{figure*}[htp!]
	\centering
	{\includegraphics[width = .85\textwidth]{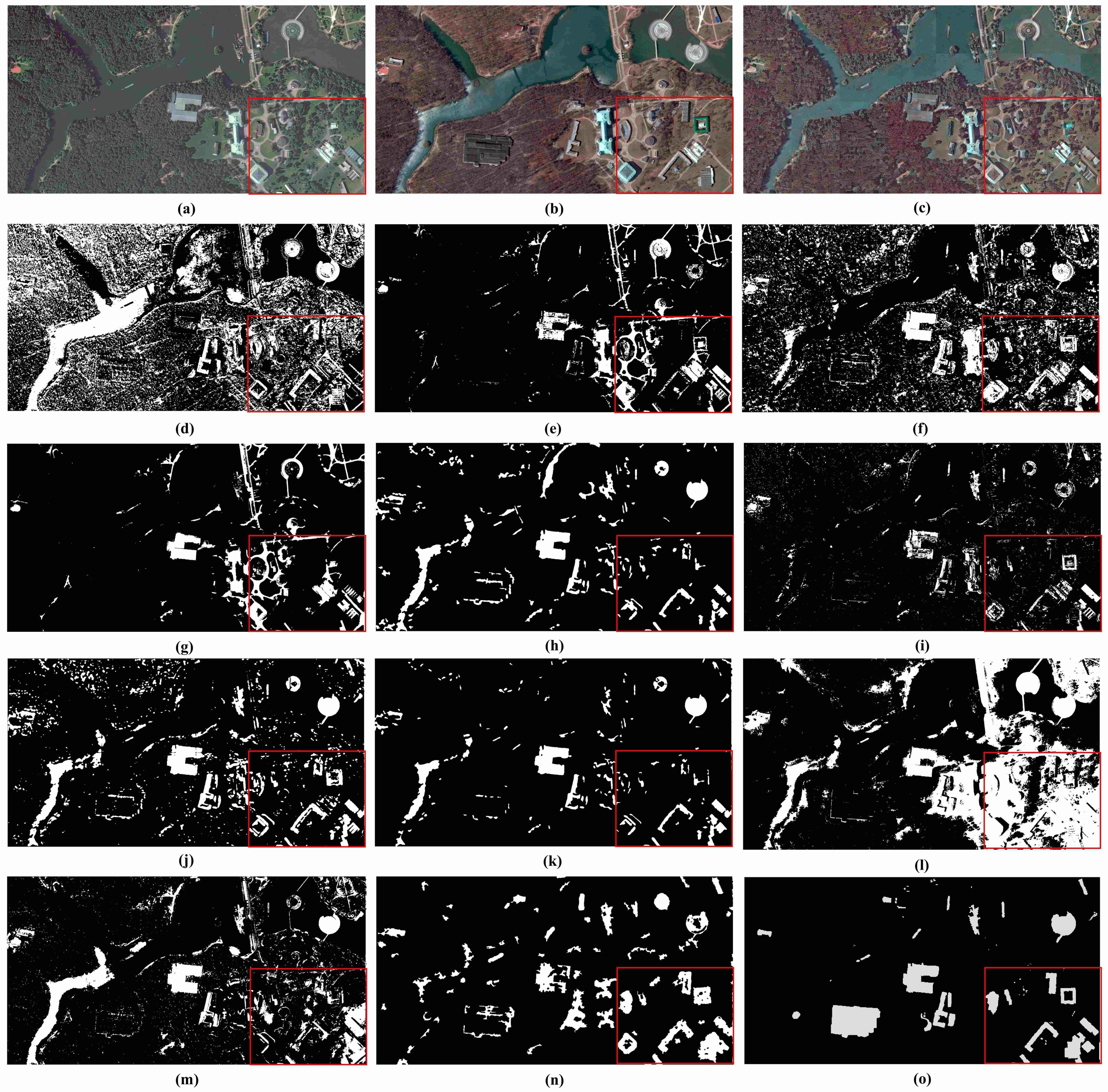}}
	\caption{Datasets and CD results of different approaches over seasonal differences from summer to autumn. The CD result is described as a binary image, in which white pixels represent change regions and black pixels represent constant regions. (a) Pre-change. (b) Post-change. (c) Translated image by UCDFormer. (d) PCA. (e) MAD. (f) IR-MAD. (g) cGAN. (h) DCVA. (i) DSFA. (j) KPCAMNet. (k) GMCD. (l) X-Net. (m) Code-Aligned Autoencoder. (n) Our proposed UCDFormer. (o) Standard reference change map. The red box denotes the representative area.}
	\label{fig:21}
\end{figure*}

\begin{figure*}[htp!]
	\centering
	{\includegraphics[width = .85\textwidth]{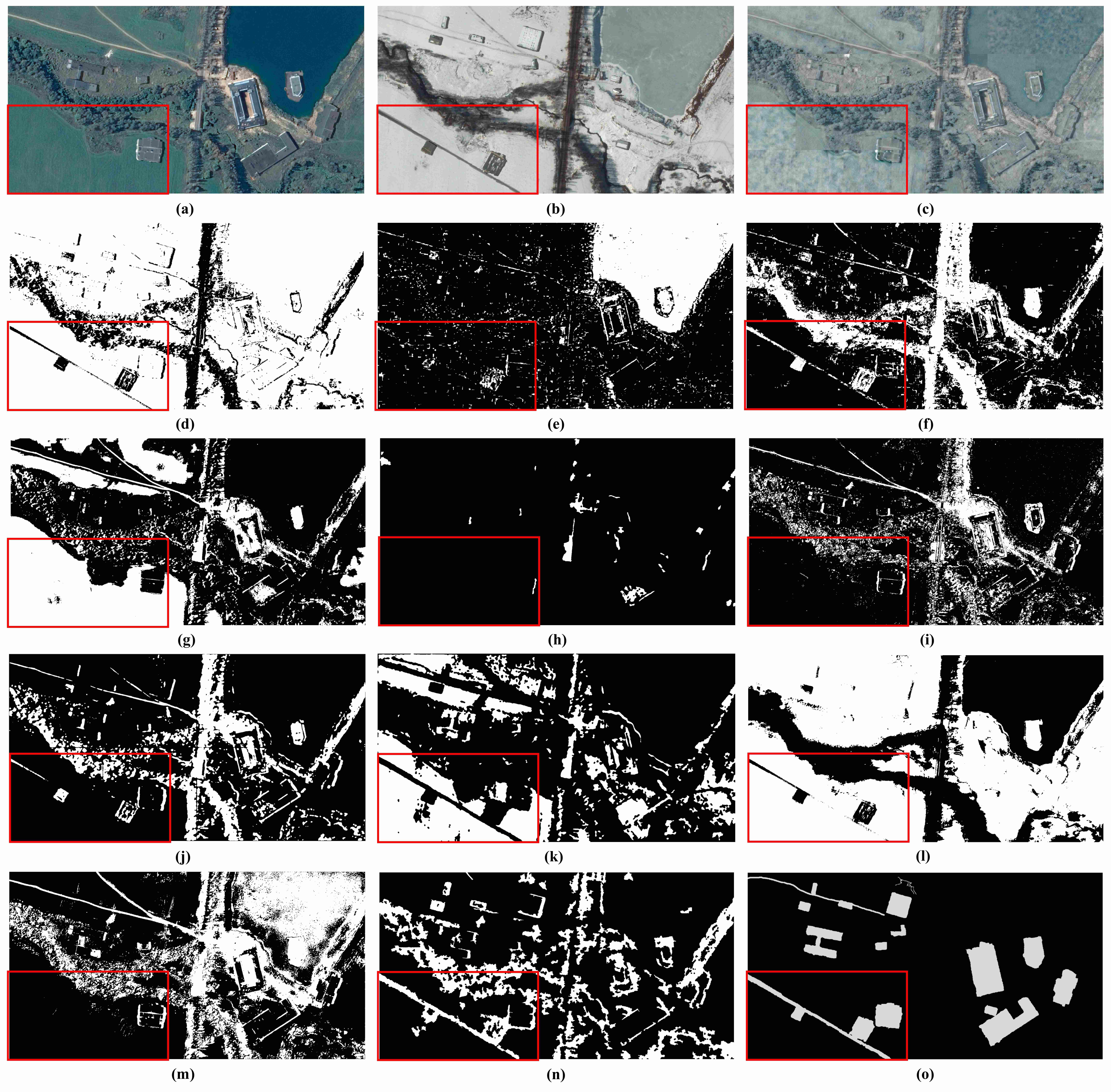}}
	\caption{Datasets and CD results of different approaches over seasonal differences from spring to winter. The CD result is described as a binary image, in which white pixels represent change regions and black pixels represent constant regions. (a) Pre-change. (b) Post-change. (c) Translated image by UCDFormer. (d) PCA. (e) MAD. (f) IR-MAD. (g) cGAN. (h) DCVA. (i) DSFA. (j) KPCAMNet. (k) GMCD. (l) X-Net. (m) Code-Aligned Autoencoder. (n) Our proposed UCDFormer. (o) Standard reference change map. The red box denotes the representative area.}
	\label{fig:22}
\end{figure*}

\begin{figure*}[htp!]
	\centering
	{\includegraphics[width = .75\textwidth]{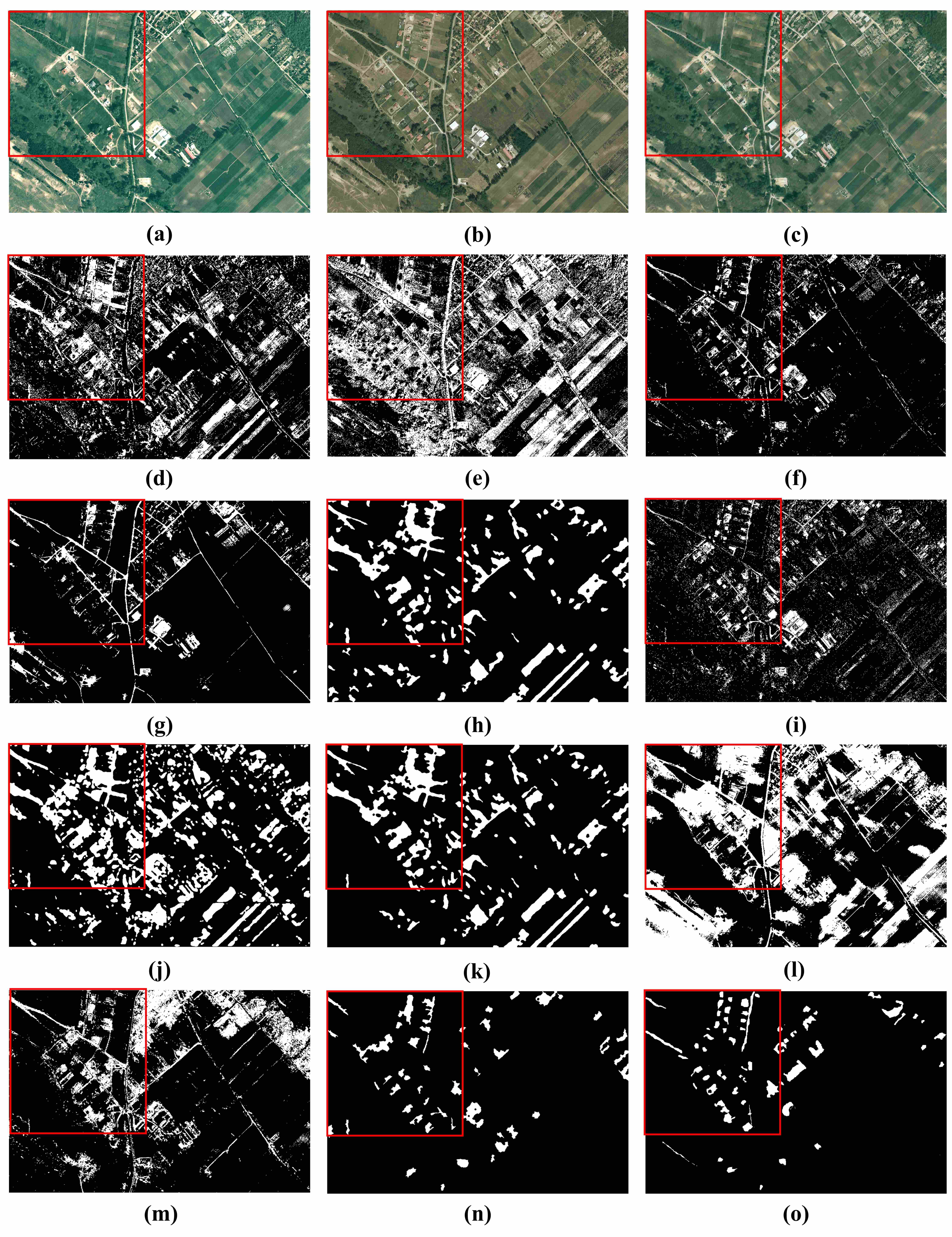}}
	\caption{Datasets and CD results of different approaches over style differences. The CD result is described as a binary image, in which white pixels represent change regions and black pixels represent constant regions. (a) Pre-change. (b) Post-change. (c) Translated image by UCDFormer. (d) PCA. (e) MAD. (f) IR-MAD. (g) cGAN. (h) DCVA. (i) DSFA. (j) KPCAMNet. (k) GMCD. (l) X-Net. (m) Code-Aligned Autoencoder. (n) Our proposed UCDFormer. (o) Standard reference change map. The red box denotes the representative area.}
	\label{fig:3}
\end{figure*}

\begin{table*}[htp!]
	\caption{Accuracy assessment on the SZTAKI dataset with  style change. $\uparrow$ represents that a higher numerical value corresponds to a better result, while $\downarrow$ represents that a lower numerical value leads to a better result.}
	\centering
	\resizebox{1.0\textwidth}{!}
	{
		\begin{tabular}{cccccccccc}
			\hline
			& \multicolumn{4}{c}{Changed}                             & \multicolumn{2}{c}{Unchanged} &                      &                      &                         \\ \cline{2-7}
			\multirow{-2}{*}{Methods}              & Recall ($\uparrow$)        & Precision ($\uparrow$)     & MDR ($\downarrow$)           & FAR  ($\downarrow$)         & Recall  ($\uparrow$)       & Precision  ($\uparrow$)    & \multirow{-2}{*}{OA ($\uparrow$)} & \multirow{-2}{*}{F1 ($\uparrow$)} & \multirow{-2}{*}{Kappa ($\uparrow$)} \\ \hline
			PCA~\cite{celik2009unsupervised}                                       & 56.16          & 10.66          & 43.84          & 19.38         & 80.62              & 97.81             & 79.65                & 17.92                & 12.07                   \\
			MAD~\cite{nielsen1998multivariate}                                     & 42.48          & 4.27           & 57.52          & 39.19         & 60.81              & 96.25             & 60.09                & 7.76                 & 0.62                    \\
			IR-MAD~\cite{nielsen2007regularized}                                 & 48.61          & 25.87          & 51.39          & 5.73          & 94.27              & 97.80             & 92.46                & 33.77                & 30.17                   \\
			cGAN~\cite{niu2018conditional}                                    & 33.90          & 16.08          & 66.10          & 7.29          & 92.71              & 97.14             & 90.38                & 21.82                & 17.38                   \\
			DCVA~\cite{saha2019unsupervised}                                    & 61.70          & 18.13          & 38.30          & 11.47         & 88.53              & 98.25             & 87.47                & 28.02                & 23.34                   \\
			DSFA~\cite{du2019unsupervised}                                    & 50.43          & 18.82          & 49.57          & 8.95          & 91.05              & 97.81             & 89.44                & 27.42                & 22.98                   \\
			KPCAMNet~\cite{wu2021unsupervised}                                & 81.62          & 14.31          & 18.38          & 20.12         & 79.88              & \textbf{99.06}             & 79.95                & 24.35                & 18.89                   \\
			GMCD~\cite{tang2021unsupervised}            & 64.57          & 18.85          & 35.43          & 11.44         & 88.56              & 98.38             & 87.61                & 29.18                & 24.57                   \\
			X-Net~\cite{luppino2021deep}           & \textbf{84.36} & 8.53           & \textbf{15.64} & 3.47          & 62.75              & 98.98             & 63.60                & 15.49                & 8.95                    \\
			Code-Aligned Autoencoder~\cite{luppino2022code}              & 62.85          & 15.68          & 37.15          & 13.92         & 86.08              & 98.25             & 85.17                & 25.10                & 20.03                   \\ \hline
			UCDFormer                              & 40.60          & \textbf{39.10}          & 59.40          & \textbf{2.60}          & \textbf{97.40}              & 97.55             & \textbf{95.15}                & \textbf{39.84}       & \textbf{37.31}          \\ \hline
	\end{tabular}}
	\label{Tab:Table3}
\end{table*}

\begin{figure*}[htp!]
	\centering
	{\includegraphics[width = .8\textwidth]{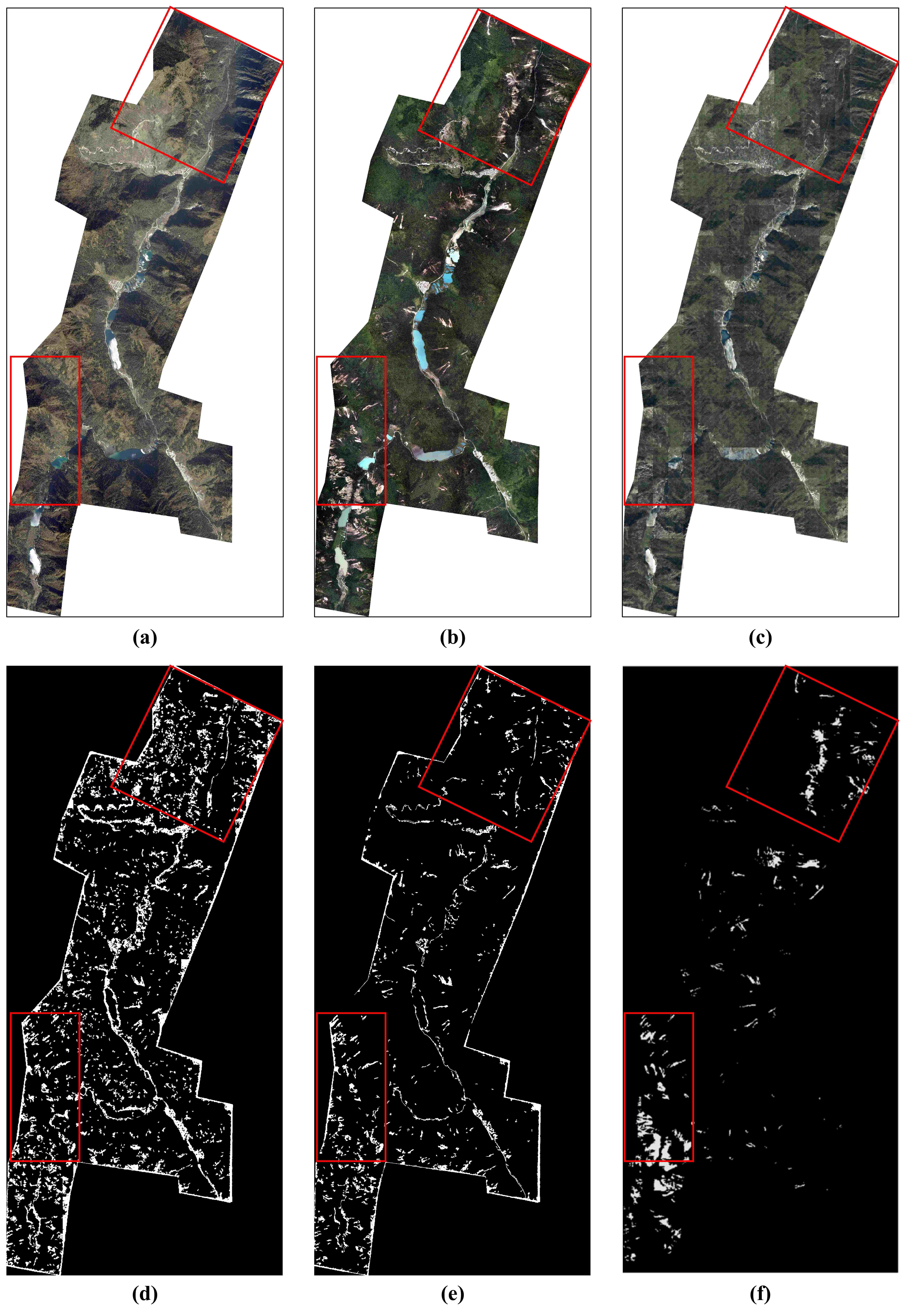}}
	\caption{Earthquake-induced landslide dataset with style differences. The CD result is described as a binary image, in which white pixels represent change regions and black pixels represent constant regions. (a) Pre-earthquake. (b) Post-earthquake. (c) Translated image by UCDFormer. (d) UCDFormer without transformer-driven image translation. (e) Our proposed UCDFormer. (f) Standard reference change map. The red box denotes the representative area.}
	\label{fig:4}
\end{figure*}

\begin{table*}[htp!]
	\caption{Accuracy assessment on the landslide detection with  style change. $\uparrow$ represents that a higher numerical value corresponds to a better result, while $\downarrow$ represents that a lower numerical value leads to a better result.}
	\centering
	\resizebox{1.0\textwidth}{!}
	{
	\begin{tabular}{ccccccccccc}
		\hline
		\multirow{2}{*}{Methods}                              & \multirow{2}{*}{Parameters of the self-attention module} & \multicolumn{4}{c}{Changed}                                      & \multicolumn{2}{c}{Unchanged}   & \multirow{2}{*}{OA ($\uparrow$)} & \multirow{2}{*}{F1 ($\uparrow$)} & \multirow{2}{*}{Kappa ($\uparrow$)} \\ \cline{3-8}
		&                                                            & Recall ($\uparrow$)         & Precision ($\uparrow$)     & MDR  ($\downarrow$)           & FAR  ($\downarrow$)         & Recall ($\uparrow$)        & Precision  ($\uparrow$)    &                     &                     &                        \\ \hline
		UCDFormer w/ Traditional Self-attention (Non Light-weight)               & 67.1M                                                   & 4.45           & 0.89           & 95.55          & 9.74          & 90.26          & 97.96          & 88.61               & 1.49                & 1.79                   \\
		UCDFormer w/o Affinity   Weight  in Image Translation & 10.7M                                                  & 41.10          & 24.39          & 58.90          & 2.51          & 97.49          & 98.83          & \textbf{96.41}      & 30.62               & 28.89                  \\
		UCDFormer w/o Image   Translation                     & -                                                          & 56.58          & 13.80          & 43.42          & 6.95          & 93.05          & 99.09          & 92.34               & 22.18               & 19.69                  \\
		UCDFormer w/ Adaptive   Threshold Only                & 10.7M                                                   & 1.66           & 3.09           & 98.34          & \textbf{1.02} & \textbf{98.98} & 98.08          & 97.10               & 2.16                & 0.82                   \\
		UCDFormer w/ FCM Only                                 & 10.7M                                                   & \textbf{63.73} & 7.61           & \textbf{36.27} & 15.22         & 84.78          & \textbf{99.17} & 84.38               & 13.60               & 10.51                  \\ \hline
		UCDFormer                                             & 10.7M                                                   & 48.04          & \textbf{25.00} & 51.96          & 2.83          & 97.17          & 98.96          & 96.22               & \textbf{32.89}      & \textbf{31.14}         \\ \hline
\end{tabular}}
\label{Tab:Table4}
\end{table*}

\subsubsection{Seasonal Change from Summer to Autumn}
We report numerical results of seasonal change from summer to autumn in Table~\ref{Tab:Table1}. The proposed UCDFormer achieves much better performance than all the other baselines. Specifically, increases in Kappa are 
47.44\%, 30.93\%, 22.18\%, 36.58\%, 12.50\%, 22.26\%, 12.52\%, 7.04\%, 35.80\%, and 20.91\% compared with  PCA~\cite{celik2009unsupervised}, MAD~\cite{nielsen1998multivariate}, IR-MAD~\cite{nielsen2007regularized}, cGAN~\cite{niu2018conditional},  DCVA~\cite{saha2019unsupervised}, DSFA~\cite{du2019unsupervised},  KPCAMNet~\cite{wu2021unsupervised}, GMCD~\cite{tang2021unsupervised}, X-Net~\cite{luppino2021deep},  and Code-Aligned Autoencoder~\cite{luppino2022code}, respectively. 
This significant gain is mainly due to the high quality of the translated images and the robust selection strategy of reliable changed and unchanged pixels.

Fig.~\ref{fig:21} shows the CD results obtained by these comparison methods. As shown in Fig.~\ref{fig:21}(c), it can be clearly observed that the  content of the translated image is coherent with the pre-change image (Fig.~\ref{fig:21}(a)). 
Furthermore,  the styles of unchanged areas in  the translated image are closer to those of the post-change image (Fig.~\ref{fig:21}(b)) and the styles of changed areas are closer to those of the pre-change image (Fig.~\ref{fig:21}(a)), which in practice means better discrimination between changed and unchanged classes.  
 We take the region in the red box of Fig.~\ref{fig:21} as an example. 
 Compared with the standard reference change map (Fig.~\ref{fig:21}(o)), PCA (Fig.~\ref{fig:21}(d)), MAD (Fig.~\ref{fig:21}(e)), cGAN (Fig.~\ref{fig:21}(g)),  DCVA (Fig.~\ref{fig:21}(h)), DSFA (Fig.~\ref{fig:21}(i)), GMCD (Fig.~\ref{fig:21}(k)), and Code-Aligned Autoencoder (Fig.~\ref{fig:21}(m))
   cannot  effectively identify some change areas due to the lack of distinguishability in the image or feature space, which results in high MDR. In addition, IR-MAD (Fig.~\ref{fig:21}(f))
 KPCAMNet (Fig.~\ref{fig:21}(j)) cannot separate various changes well, resulting in a great deal of noise in the detection results. Importantly, there are numerous incorrectly changed regions in the cGAN (Fig.~\ref{fig:21}(g)), X-Net (Fig.~\ref{fig:21}(l)), and Code-Aligned Autoencoder (Fig.~\ref{fig:21}(m)) results due to inappropriate image translation methods (such as GAN) for change detection with seasonal differences. In contrast, Fig.~\ref{fig:21}(n) shows the change map of the proposed method. It is worth noting that an excellent visualized result of the change map is obtained by the joint use of transformer-driven image translation and reliable change extraction.
\subsubsection{Seasonal Change from Spring to Winter}
The quantitative comparison experimental results for unsupervised CD from spring to winter are shown in Table~\ref{Tab:Table2}. Similar to the performance on the seasonal change from summer to autumn, the proposed UCDFormer outperforms all the other methods and achieves the highest Kappa value. Specifically, increments of 21.15\%, 20.01\%, 16.20\%, 15.47\%,  17.22\%, 14.11\%, 14.35\%, 18.96\%, 13.32\%, and 19.80\% in Kappa are obtained compared with PCA~\cite{celik2009unsupervised}, MAD~\cite{nielsen1998multivariate}, IR-MAD~\cite{nielsen2007regularized}, cGAN~\cite{niu2018conditional},  DCVA~\cite{saha2019unsupervised}, DSFA~\cite{du2019unsupervised},  KPCAMNet~\cite{wu2021unsupervised}, GMCD~\cite{tang2021unsupervised}, X-Net~\cite{luppino2021deep},  and Code-Aligned Autoencoder~\cite{luppino2022code}, respectively. Again, the effectiveness and robustness of UCDFormer are verified.

Fig.~\ref{fig:22} shows the translated image and  binary change maps of PCA, MAD, IR-MAD, cGAN,  DCVA, DSFA,  KPCAMNet, GMCD, X-Net, Code-Aligned Autoencoder, and our proposed UCDFormer. 
Unsupervised CD on seasonal change from spring to winter is  more challenging  than seasonal change from summer to autumn, due to the larger domain shift between the images from spring to winter. However, after translating Fig.~\ref{fig:22}(a) into Fig.~\ref{fig:22}(c) by employing the proposed transformer-driven image translation module, the domain shift between Fig.~\ref{fig:22}(a) and Fig.~\ref{fig:22}(b) can be effectively reduced for change extraction of UCDFormer. Furthermore, the performance trend in the visualization results of these comparable methods is similar to Table~\ref{Tab:Table2}.
 Specifically, in the binary change maps generated by PCA (Fig.~\ref{fig:22}(d)),  DCVA (Fig.~\ref{fig:22}(h)), DSFA (Fig.~\ref{fig:22}(i)),  GMCD (Fig.~\ref{fig:22}(k)), X-Net (Fig.~\ref{fig:22}(l)), and Code-Aligned Autoencoder (Fig.~\ref{fig:22}(m)), both of the new building and road in the red box of Fig.~\ref{fig:22} are not detected.
 MAD (Fig.~\ref{fig:22}(e)) and KPCAMNet (Fig.~\ref{fig:22}(j)) detect the new building areas, while the new road areas are still not detected. For cGAN (Fig.~\ref{fig:22}(g)) and X-Net (Fig.~\ref{fig:22}(l)), most of the changed regions are  incorrect due to the inappropriate GAN-based image translation network for homogeneous remote sensing images. IR-MAD (Fig.~\ref{fig:22}(f)) exhibits better visualization effects for change targets. 
 Compared with the above methods, the proposed UCDFormer is able to achieve a  more accurate change map, as shown in Fig.~\ref{fig:22}(n).

\subsection{Experiments on Style Changes}

\subsubsection{SZTAKI Data with Style Differences}
The style differences between the image pair from SZTAKI are caused by differences in light conditions or atmospheric conditions. Table~\ref{Tab:Table3} shows the quantitative comparison of all methods on unsupervised CD with style differences, in which the proposed UCDFormer outperforms other comparative methods. That is, UCDFormer achieves  higher OA, F1, and Kappa. For example, compared with PCA~\cite{celik2009unsupervised}, MAD~\cite{nielsen1998multivariate}, IR-MAD~\cite{nielsen2007regularized}, cGAN~\cite{niu2018conditional},  DCVA~\cite{saha2019unsupervised}, DSFA~\cite{du2019unsupervised},  KPCAMNet~\cite{wu2021unsupervised}, GMCD~\cite{tang2021unsupervised}, X-Net~\cite{luppino2021deep},  and Code-Aligned Autoencoder~\cite{luppino2022code}, increases in Kappa are 25.24\%, 36.69\%, 7.14\%, 19.93\%, 13.97\%, 14.33\%, 18.42\%, 12.74\%, 28.36\%, and 17.28\%, respectively. This demonstrates that the proposed UCDFormer is effective for change detection with style differences.

Fig.~\ref{fig:3} shows the translated image and  binary change maps of  PCA, MAD, IR-MAD, cGAN,  DCVA, DSFA,  KPCAMNet, GMCD, X-Net, Code-Aligned Autoencoder, and the proposed UCDFormer on the SZTAKI dataset with style differences. As shown in  Fig.~\ref{fig:3}(c), to improve the distinguishability of change areas, the style of the translated  image is between the pre-change image (Fig.~\ref{fig:3}(a)) and post-change image (Fig.~\ref{fig:3}(b)) by using the proposed transformer-driven image translation. 
Furthermore, as shown in these binary change maps, UCDFormer detects changes successfully in all comparable methods. There are too many misclassified pixels and too much noise in  PCA, MAD, IR-MAD, cGAN,  DCVA, DSFA,  KPCAMNet, GMCD, X-Net, and Code-Aligned Autoencoder, due to the domain shift caused by style differences, as shown in Fig.~\ref{fig:3}(d)-(l) and (m), respectively. 
More specifically,  the farmland and road in the red box of Fig.~\ref{fig:3} are classified as changed regions due to the substantial style differences. However, the proposed UCDFormer (Fig.~\ref{fig:3}(n)) overcomes the style differences and classifies the farmland and road properly. In addition, compared with the standard reference change map (Fig.~\ref{fig:3}(o)), it can be found that  new buildings are missed in Fig.~\ref{fig:3}(n), probably because the change in this region is too subtle to be detected.

\begin{figure*}[htp!]
	\centering
	{\includegraphics[width = .8\textwidth]{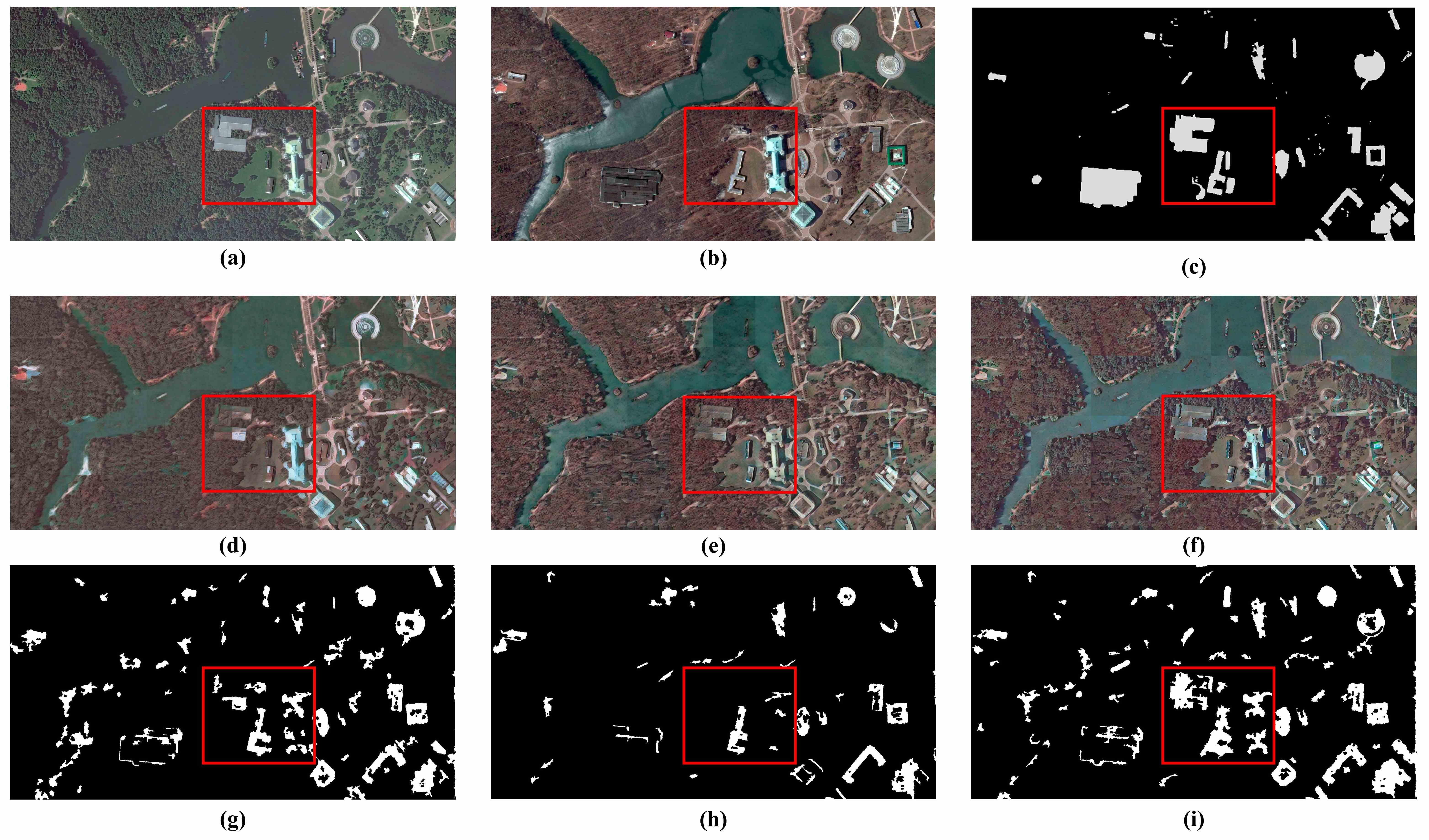}}
	\caption{Comparison of different transformer-based image translation methods. (a) Pre-change. (b) Post-change. (c) Standard reference change map. (d) Translated image by StyTr2. (e) Translated image by Styleformer. (f) Translated image by UCDFormer. (g) Binary result of StyTr2-based change detection. (h) Binary result of  Styleformer-based change detection. (i) UCDFormer. The red box denotes the representative area.}
	\label{fig:5}
\end{figure*}

\begin{table*}[tp!]
	\caption{Comparison results of different transformer-based image translation methods. $\uparrow$ represents that a higher numerical value corresponds to a better result, while $\downarrow$ represents that a lower numerical value leads to a better result.}
	\centering
	\resizebox{1.0\textwidth}{!}
	{\begin{tabular}{ccccccccccc}
			\hline
			\multirow{2}{*}{Methods} & \multirow{2}{*}{Inference time for  resolution 256$\times$256  (in seconds)} & \multicolumn{4}{c}{Changed}                                      & \multicolumn{2}{c}{Unchanged}   & \multirow{2}{*}{OA ($\uparrow$)} & \multirow{2}{*}{F1 ($\uparrow$)} & \multirow{2}{*}{Kappa ($\uparrow$)} \\ \cline{3-8}
		&                                                            & Recall ($\uparrow$)         & Precision ($\uparrow$)     & MDR  ($\downarrow$)           & FAR  ($\downarrow$)         & Recall ($\uparrow$)        & Precision  ($\uparrow$)    &                     &                     &                        \\ \hline
			Styleformer~\cite{wu2021styleformer}              & 0.029                                                                  & 31.71          & \textbf{54.22} & 68.29          & \textbf{2.18} & \textbf{97.82} & 94.63          & \textbf{92.85}               & 40.01                        & 36.49                  \\
			StyTr2~\cite{deng2022stytr2}                    & 0.136                                                                  & 49.95          & 45.16          & 50.05          & 4.93          & 95.07          & 95.89          & 91.67                        & 47.43                        & 42.92                  \\ \hline
			UCDFormer                & 0.031                                                                  & \textbf{59.10} & 50.21          & \textbf{40.90} & 4.77          & 95.23          & \textbf{96.62} & 92.52                        & \textbf{54.30}               & \textbf{50.25}         \\ \hline
	\end{tabular}}
	\label{Tab:Table5}
\end{table*}

\begin{table*}[tp!]
	\caption{Ablation study of image translation and change extraction on different datasets. $\uparrow$ represents that a higher numerical value corresponds to a better result, while $\downarrow$ represents that a lower numerical value leads to a better result.}
	\centering
	\resizebox{1.0\textwidth}{!}
	{\begin{tabular}{ccccccccccc}
\hline
\multirow{2}{*}{Datasets}                                & \multirow{2}{*}{Methods}      & \multicolumn{4}{c}{Changed}                                      & \multicolumn{2}{c}{Unchanged}   & \multirow{2}{*}{OA ($\uparrow$)} & \multirow{2}{*}{F1 ($\uparrow$)} & \multirow{2}{*}{Kappa ($\uparrow$)} \\ \cline{3-8}
		&                                                            & Recall ($\uparrow$)         & Precision ($\uparrow$)     & MDR  ($\downarrow$)           & FAR  ($\downarrow$)         & Recall ($\uparrow$)        & Precision  ($\uparrow$)    &                     &                     &                        \\ \hline
\multirow{4}{*}{\makecell[c]{Seasonal change from  \\ summer to autumn}} & UCDFormer w/o Image Translation        & 64.77           & 42.34              & 35.23          & 7.17          & 92.83            & 97.01               & 90.72                        & 51.21                        & 46.33                           \\
                                                                  & UCDFormer w/ Adaptive   Threshold Only & 5.64            & \textbf{70.00}     & 94.36          & \textbf{0.20} & \textbf{99.80}   & 92.86               & \textbf{92.72}               & 10.44                        & 9.43                            \\
                                                                  & UCDFormer w/ FCM Only                  & \textbf{78.62}  & 23.47              & \textbf{21.38} & 20.86         & 79.14            & \textbf{97.85}      & 79.10                        & 36.14                        & 27.78                           \\ \cline{2-11} 
                                                                  & UCDFormer                              & 59.10           & 50.21              & 40.90          & 4.77          & 95.23            & 96.62               & 92.52                        & \textbf{54.30}               & \textbf{50.25}                  \\ \hline
\multirow{4}{*}{\makecell[c]{Seasonal change from \\  spring to winter}} & UCDFormer w/o Image Translation        & 29.27           & 21.96              & 70.73          & 12.69         & 87.31            & 91.01               & 81.00                        & 25.10                        & 14.47                           \\
                                                                  & UCDFormer w/ Adaptive   Threshold Only & 1.31            & \textbf{42.49}     & 98.69          & \textbf{0.22} & \textbf{99.78}   & 89.23               & \textbf{89.08}               & 2.55                         & 1.91                            \\
                                                                  & UCDFormer w/ FCM Only                  & \textbf{55.67}  & 23.41              & \textbf{44.33} & 22.22         & 77.78            & \textbf{93.50}      & 75.37                        & \textbf{32.96}               & 20.84                           \\ \cline{2-11} 
                                                                  & UCDFormer                              & 47.45           & 25.07              & 52.55          & 17.30         & 82.70            & 92.80               & 78.86                        & 32.80                        & \textbf{21.65}                  \\ \hline
\multirow{4}{*}{\makecell[c]{SZTAKI dataset  \\ with  style change}}     & UCDFormer w/o Image Translation        & 33.89           & 34.89              & 66.11          & 2.60          & 97.40            & 97.28               & 94.89                        & 34.38                        & 31.72                           \\
                                                                  & UCDFormer w/ Adaptive   Threshold Only & 1.03            & \textbf{61.18}     & 98.97          & \textbf{0.03} & \textbf{99.97}   & 96.08               & \textbf{96.06}               & 2.03                         & 1.90                            \\
                                                                  & UCDFormer w/ FCM Only                  & \textbf{82.73}  & 15.60              & \textbf{17.27} & 18.43         & 81.57            & \textbf{99.14}      & 81.61                        & 26.24                        & 20.99                           \\ \cline{2-11} 
                                                                  & UCDFormer                              & 40.60           & 39.10              & 59.40          & 2.60          & 97.40            & 97.55               & 95.15                        & \textbf{39.84}               & \textbf{37.31}                  \\ \hline
\end{tabular}}
\label{Tab:Table60}
\end{table*}

\begin{table*}[tp!]
	\caption{Analysis of Hyperparameter ($\gamma$) in loss function. 
$\uparrow$ represents that a higher numerical value corresponds to a better result, while $\downarrow$ represents that a lower numerical value leads to a better result.}
	\centering
	\resizebox{1.0\textwidth}{!}
	{\begin{tabular}{cccccccccccc}
\hline
Datasets                                               & $\gamma$ & 1     & 100   & 300   & 500            & 800   & 1000           & 1300  & 1500           & 1800  & 2000  \\ \hline
\multirow{3}{*}{Seasonal change from summer to autumn} & OA ($\uparrow$)       & 81.39 & 93.31 & 93.64 & 93.56          & 93.99 & 92.52          & \textbf{94.13} & 94.09 & 94.04 & 94.04 \\
                                                       & F1 ($\uparrow$)       & 37.90 & 35.66 & 40.64 & 39.43          & 46.43 & \textbf{54.30} & 47.77 & 47.82          & 46.85 & 46.11 \\
                                                       & Kappa ($\uparrow$)    & 30.02 & 32.86 & 37.86 & 36.64          & 43.64 & \textbf{50.25} & 45.04 & 45.05          & 44.08 & 43.39 \\ \hline
\multirow{3}{*}{SZTAKI dataset with  style   change}   & OA ($\uparrow$)      & 95.75 & 95.44 & 95.21 & \textbf{95.77} & 94.74 & 95.15          & 95.30 & 94.85          & 94.94 & 95.05 \\
                                                       & F1 ($\uparrow$)      & 29.10 & 30.54 & 31.86 & 36.7           & 36.80 & \textbf{39.84} & 38.87 & 35.14          & 34.19 & 33.91 \\
                                                       & Kappa ($\uparrow$)    & 27.14 & 28.28 & 29.42 & 34.58          & 34.07 & \textbf{37.31} & 36.43 & 32.46          & 31.56 & 31.34 \\ \hline
\end{tabular}}
\label{Tab:Table61}
\end{table*}

\subsubsection{Applications in Landslide Detection with Style Differences}
The proposed UCDFormer is conducted for large-scale earthquake-induced landslide detection with style differences. To prove the necessity of the transformer-driven image translation and the change extraction stages when considering large-scale applications, ablation experiments are conducted, which is shown in Table~\ref{Tab:Table4}. 
It can be seen that UCDFormer achieves excellent performance for the large-scale landslide detection task. OA, F1, and Kappa have reached 96.22\%, 32.89\%, and 31.14\%, respectively. First, for the transformer-driven image translation stage, by comparing UCDFormer and UCDFormer without image translation, the proposed transformer-driven image translation module in UCDFormer increases the performance in Kappa by 11.45\%, which represents a significant gain. Furthermore, the two core modules (efficient self-attention and affinity weight) in the transformer-driven image translation stage are analyzed, respectively.  In order to verify the effectiveness and practicality of the proposed efficient self-attention module, the traditional multi-head self-attention layer (Eq.~\ref{eq1}) in  the regular transformer~\cite{vaswani2017attention, dosovitskiy2020image} is used to replace the efficient self-attention module in UCDFormer, called UCDFormer with traditional self-attention), and serves as a benchmark comparison method for self-attention modules. We can observe that the proposed efficient self-attention module achieves a significant performance gain while reducing    the parameters of the self-attention module  by a factor of approximately six, by comparing UCDFormer and UCDFormer with traditional self-attention.  In addition, by comparing UCDFormer and UCDFormer without affinity   weight  in image translation, it has been demonstrated that the proposed affinity   weight is effective for improving the performance of change maps (increments of 2.25 \% in Kappa). 
Second, in order to verify the efficacy of the proposed change extraction module, we perform an ablation study that evaluates variants of UCDFormer.  As shown in Table~\ref{Tab:Table4}, UCDFormer outperforms UCDFormer w/ adaptive threshold only and UCDFormer w/ FCM only, which indicates that the proposed change extraction is effective and necessary for the CD with domain shift tasks. 

Fig.~\ref{fig:4} shows the translated image and  binary change maps of UCDFormer without image translation and UCDFormer with image translation. The binary change map of UCDFormer without image translation (Fig.~\ref{fig:4}(d)) is obtained by applying the proposed change extraction module directly to the pre-earthquake image (Fig.~\ref{fig:4}(a)) and post-earthquake image (Fig.~\ref{fig:4}(b)). In addition, the binary change map of UCDFormer  with image translation (Fig.~\ref{fig:4}(e))  is obtained by performing the  change extraction module on the translated image (Fig.~\ref{fig:4}(c)) and post-earthquake image (Fig.~\ref{fig:4}(b)). By comparing these two results (especially the red box area in Fig.~\ref{fig:4}), it can be seen that the style differences cause many noise points and a large number of misidentified landslide areas (such as roads, and shadow areas) in the binary change map. However, the proposed transformer-driven image translation can alleviate these misidentified areas and noise effectively, thereby greatly improving the final binary results. 

\subsection{Model Analysis}
\subsubsection{Hyperparameter Analysis in Image Translation}
The hyperparameter $\gamma$ in the loss function (Eq.~\ref{eq.10}) is for balancing the loss. We have undertaken a thorough analysis of this hyperparameter using  the dataset with seasonal differences from summer to autumn and the SZTAKI dataset with  style   change, which is shown in Table~\ref{Tab:Table61}. 
We have observed that the loss of the content loss is 1000 times smaller compared to the weighted translation loss when outputting the loss values. Thus, we perform a sensitivity analysis on the hyperparameter $\gamma$ within the range of 1 to 2000. As shown in Table~\ref{Tab:Table61}, we can clearly observe that different $\gamma$ has a great impact on the model results. 
Importantly, when $\gamma$=1000, the model performances (Kappa) are the best. Thus, the $\gamma$ is set to 1000 by default in our different experiments.
\subsubsection{Comparison of Different Transformer-based Image Translation Methods}
In order to evaluate the effectiveness of the proposed light-weight transformer-based image translation, two mainstream transformer-based image translation methods, including Styleformer~\cite{wu2021styleformer} and StyTr2~\cite{deng2022stytr2}, are utilized for comparison.  The results are presented in Table~\ref{Tab:Table5}. It can be seen that our proposed method exhibits better performance than these two transformer-based methods. In addition, we compared the inference time of different transformer-based methods for an output image with a 256 $\times$ 256  resolution using a single Tesla P100 GPU, and it demonstrates the efficient and real-time image transformation capability of our method. 

Fig.~\ref{fig:5} shows the visual results of qualitative comparisons. By comparing  Fig.~\ref{fig:5}(f) with Fig.~\ref{fig:5}(d) and Fig.~\ref{fig:5}(e), UCDFormer can realize that the style of changed targets is closer to the pre-change image (content image), and the style of unchanged targets is closer to the post-change image (style image), thus improving the performance of the model. It has been demonstrated that the domain-specific affinity weight in the image translation stage is effective and practical for unsupervised CD.
\subsubsection{Ablation Study of Image Translation and Change Extraction}
Ablation experiments of the image translation stage and the change extraction stage are  conducted to  investigate the effectiveness and practicality of transformer-driven image translation and change map extraction.  The results on the seasonal difference from summer to autumn, the seasonal difference  from spring to winter, and the SZTAKI dataset with  style change are presented in Table~\ref{Tab:Table60}. It is obvious that the performance of UCDFormer is better than that of UCDFormer without image translation. Again, the effectiveness and necessity of the transformer-driven image translation stage for the unsupervised CD task with domain shift  have been demonstrated. In addition, Our proposed
UCDFormer performs better than UCDFormer with  adaptive threshold only and UCDFormer with FCM only, 
indicating both the adaptive threshold and the FCM clustering is  crucial and necessary for the proposed change map extraction. Furthermore, we can observe that the performance of UCDFormer with FCM only is better than UCDFormer with  adaptive threshold only. Thus, FCM clustering is a useful unsupervised binarization tool for unsupervised CD tasks with domain shift.
\section{Conclusions  \label{co}}
This article studies unsupervised CD with domain shift, which takes into account style differences and seasonal differences between two images. In order to provide a single generalized guide for detecting the changes of remote sensing images with domain shift, we propose the UCDFormer model. In UCDFormer,  a transformer-driven image translation  is first presented to  produce visually plausible stylization images for arbitrary attribute transfer with real-time efficiency. After image translation, the domain shift due to style differences and  seasonal differences for homogeneous data can be significantly. Then, based on the difference map,  the reliable changed and unchanged pixel positions are further extracted by fusing the pseudo-change maps of FCM clustering and adaptive threshold.  
Finally, the RF classifier is employed to perform the binary classification task, which is trained on the selected changed and unchanged pixel pairs. The other pixel pairs are subsequently classified by this classifier to obtain the final CD results. 

Experiments demonstrate that the proposed method achieves excellent performance when processing multi-temporal images with style differences and seasonal differences. Furthermore, the effectiveness of the proposed UCDFormer on the large-scale seismic landslide detection task proves that the generalized guidance presented has broad applications for an unsupervised CD with domain shift. In the future, we will extend the applications of the proposed method to some unsupervised CD tasks with large domain shifts and unsupervised CD for time series data.

\bibliographystyle{IEEEtran}
\bibliography{ref}

\end{document}